\documentclass{article}

\usepackage{microtype}
\usepackage{graphicx}
\usepackage{subfigure}
\usepackage{booktabs} 
\PassOptionsToPackage{numbers}{natbib}

\usepackage{hyperref}

\usepackage[accepted]{icml2025} 

\usepackage{amsmath}
\usepackage{amssymb}
\usepackage{mathtools}
\usepackage{amsthm}

\usepackage[capitalize,noabbrev]{cleveref}

\theoremstyle{plain}

\theoremstyle{definition}

\theoremstyle{remark}

\usepackage[textsize=tiny]{todonotes}
\usepackage{enumitem}

\newcommand{\skipSpace}{\! \! \! \! \! \!}
\newcommand{\tmin}{t_{\text{min}}}
\newcommand{\tmax}{t_{\text{max}}}
\newcommand{\vx}{\boldsymbol{x}}
\newcommand{\vz}{\boldsymbol{z}}

\usepackage{cleveref}

\icmltitlerunning{Effective and Efficient Masked Image Generation Models}

\begin{document}

\twocolumn[
\icmltitle{Effective and Efficient Masked Image Generation Models}



\icmlsetsymbol{intern}{$\dagger$}
\icmlsetsymbol{cor}{$\mathparagraph$}

\begin{icmlauthorlist}
\icmlauthor{Zebin You}{ruc1,ruc2,ruc3,intern}
\icmlauthor{Jingyang Ou}{ruc1,ruc2,ruc3}
\icmlauthor{Xiaolu Zhang}{ant}
\icmlauthor{Jun Hu}{ant}
\icmlauthor{JUN ZHOU}{ant}
\icmlauthor{Chongxuan Li}{ruc1,ruc2,ruc3,cor}
\end{icmlauthorlist}

\icmlaffiliation{ruc1}{Gaoling School of Artificial Intelligence, Renmin University of China, Beijing, China.}
\icmlaffiliation{ruc2}{Beijing Key Laboratory of Research on Large Models and Intelligent Governance.}
\icmlaffiliation{ruc3}{Engineering Research Center of Next-Generation Intelligent Search and Recommendation, MOE.}
\icmlaffiliation{ant}{Ant Group}
\icmlcorrespondingauthor{Chongxuan Li}{chongxuanli@ruc.edu.cn}

\icmlkeywords{Effective and Efficient Masked Image Generation Models}

\vskip 0.3in
]



\printAffiliationsAndNotice{\icmlItern} 

\begin{abstract}
Although masked image generation models and masked diffusion models are designed with different motivations and objectives, we observe that they can be unified within a single framework. Building upon this insight, we carefully explore the design space of training and sampling, identifying key factors that contribute to both performance and efficiency. Based on the improvements observed during this exploration, we develop our model, referred to as \textbf{eMIGM}. Empirically, eMIGM demonstrates strong performance on ImageNet generation, as measured by Fréchet Inception Distance (FID). In particular, on ImageNet $256\times256$, with similar number of function evaluations (NFEs) and model parameters, eMIGM outperforms the seminal VAR. Moreover, as NFE and model parameters increase, eMIGM achieves performance comparable to the state-of-the-art continuous diffusion model REPA while requiring less than 45\% of the NFE. Additionally, on ImageNet $512\times512$, eMIGM outperforms the strong continuous diffusion model EDM2. Code is available at \url{https://github.com/ML-GSAI/eMIGM}.
\end{abstract}

\section{Introduction}
Masked modeling has proven effective across various domains, including self-supervised learning~\cite{he2022masked, bao2021beit, devlin2018bert}, label to image generation~\cite{li2023mage, chang2022maskgit, li2024autoregressive, ni2024adanat}, text to image generation~\cite{bai2024meissonic, shao2024bag} and text generation~\cite{sahoo2024simple, shi2024simplified, lou2024discrete}. In image generation, MaskGIT~\cite{chang2022maskgit} introduced masked image generation, offering efficiency and quality improvements over autoregressive models but still lagging behind diffusion models~\cite{ho2020denoising, sohl2015deep, song2020score} due to information loss from discrete tokenization~\cite{esser2021taming, van2017neural}. MAR~\cite{li2024autoregressive} eliminated this bottleneck via diffusion loss, achieving strong results, yet key factors (e.g., masking schedule, loss function) remain underexplored. Moreover, with limited sampling steps (e.g., 16), its performance falls short of coarse-to-fine next-scale prediction model VAR~\cite{tian2024visual}.

In parallel, masked diffusion models~\cite{sahoo2024simple, shi2024simplified, lou2024discrete, ou2024your} have shown promise in text generation, demonstrating scaling properties~\cite{nie2024scaling} similar to ARMs and offering a principled probabilistic framework for training and inference. However, their applicability to image generation remains an open question.

\begin{figure*}[!t]
    \centering
    \includegraphics[width=0.85\linewidth]{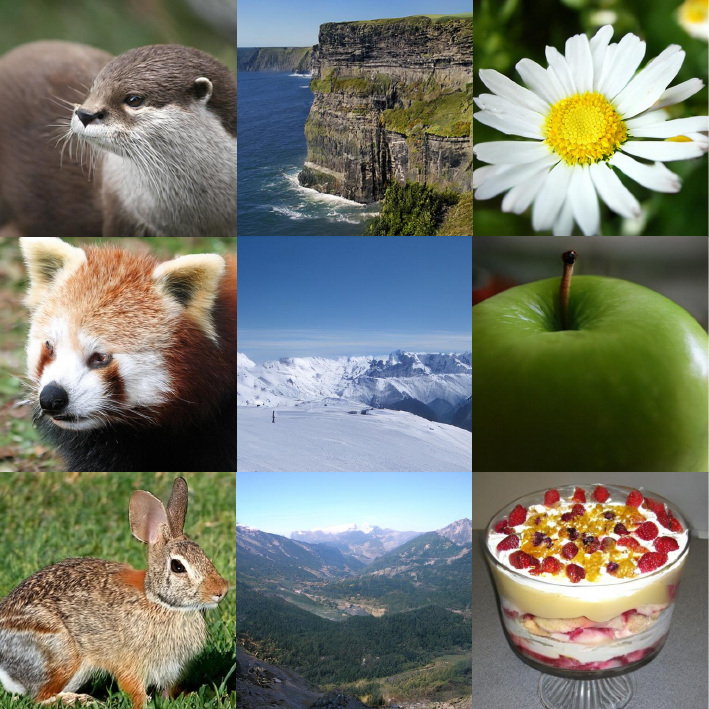}
    \caption{Generated samples from eMIGM trained on ImageNet $512\times512$.\label{fig:generated_image}}
\end{figure*}
We propose a unified framework integrating masked image modeling~\cite{chang2022maskgit, li2024autoregressive, bai2024meissonic} and masked diffusion models~\cite{lou2024discrete, sahoo2024simple, shi2024simplified}, leveraging the strengths of both paradigms. This enables a systematic exploration of training and sampling strategies to optimize performance. For training, we find that images, due to their high redundancy, benefit from a higher masking ratio, a simple weighting function inspired by MaskGIT and MAE~\cite{he2022masked} tricks, improving generation quality. We also present CFG with Mask, replacing the fake class token with a mask token for unconditional generation, further enhancing performance. For sampling, predicting fewer tokens in early stages improves results. However, early-stage guidance decreases variance, raising FID. To counter this, we propose a time interval strategy for classifier-free guidance in masked image generation, applying guidance only in later stages. This maintains strong performance while significantly accelerating sampling by reducing NFEs.

Building on our training and sampling improvements, we develop eMIGM and evaluate it on ImageNet~\cite{deng2009imagenet} at $256\times256$ and $512\times512$ resolutions. As model parameters scale, eMIGM achieves progressively higher sample quality in a predictable manner (Fig.~\ref{fig:scaling_flops_fid_scatter}). Larger models further enhance efficiency, maintaining superior quality with similar training FLOPs and sampling time (Fig.\ref{fig:scaling_flops_fid}, Fig.~\ref{fig:sample_scaling}). Notably, eMIGM delivers high-quality samples with few sampling steps. On ImageNet $256\times256$, with similar NFEs and model parameters, it consistently outperforms VAR~\cite{tian2024visual}. Increasing NFE and model size, our best-performing eMIGM-H becomes comparable to state-of-the-art diffusion models like REPA~\cite{yu2024representation} (FID 1.57 vs. 1.42)—without requiring self-supervised features. On ImageNet $512\times512$, eMIGM-L surpasses EDM2~\cite{karras2024analyzing} while using a lower parameter count, demonstrating efficiency and scalability. Qualitatively, eMIGM generates realistic and diverse images (Fig.~\ref{fig:generated_image}).

In summary, our key contributions are as follows:
\begin{itemize}[leftmargin=*]
\setlength\itemindent{.5em}
\setlength{\itemsep}{-2pt}  
\item We propose a unified formulation to systematically explore the design space of masked image generation models, uncovering the role of each component.  
\item We introduce the time interval strategy for classifier-free guidance, maintaining high performance while significantly reducing sampling time.  
\item We surpass the seminar diffusion models on ImageNet $512\times512$.  
\item We demonstrate that eMIGM benefits from scaling, with larger eMIGM models achieving greater efficiency.  
\end{itemize}

\section{Preliminaries}
\subsection{Masked Image Generation}
Let \( \boldsymbol{x} = [\boldsymbol{x}^i]_{i=1}^N \) represent the discrete tokens of an image obtained via a VQ encoder~\cite{esser2021taming, van2017neural}, and let \( \text{[M]} \) denote the special mask token. We consider two seminal masked image generation methods. 

\textbf{MaskGIT} ~\cite{chang2022maskgit} first extends the concept of masked language modeling from BERT~\cite{devlin2018bert} (i.e., predicting masked tokens based on unmasked tokens) to image generation, achieving excellent performance with low sampling cost (approximately 10 sampling steps) on ImageNet~\cite{deng2009imagenet}. However, its performance degrades when the number of sampling steps increases under its default mask schedule.

During training, MaskGIT optimizes the cross entropy loss as follows. 
A ratio $r$ is sampled from $[0, 1]$, 
and based on the mask scheduling function $ \gamma_r $,
masked image $ \boldsymbol{x}_{\overline{\textbf{M}}} $ is sampled from masking distribution $ q_{\gamma_r}(\boldsymbol{x}_{\overline{\textbf{M}}}|\boldsymbol{x}) $ that randomly masks $ \lceil N \gamma_r \rceil $ tokens of $\boldsymbol{x}$ as $ \text{[M]} $. 

The loss function is then defined as:
\begin{equation}
    \mathcal{L}(\boldsymbol{x}) =  \mathbb{E}_{r\sim U[0,1]} \mathbb{E}_{q_{\gamma_r}(\boldsymbol{x}_{\overline{\textbf{M}}}|\boldsymbol{x})}\left[ \sum_{\{i|\boldsymbol{x}^i=\text{[M]}\}}\skipSpace -\log p_{\boldsymbol{\theta}}\left( \boldsymbol{x}^i \, | \, \boldsymbol{x}_{\overline{\textbf{M}}} \right)  \right].
\label{eq:maskgit_loss}
\end{equation}
\begin{table*}[t!]
    \caption{\textbf{Comparison of different masked image modeling approaches through a unified framework.} The differences among these approaches are defined by the choice of masking distribution $q(\vx_t|\vx_0)$, weighting function $w(t)$, and conditional distribution $p_{\boldsymbol{\theta}}(\vx_0^i \mid \vx_t)$.}
    \begin{center} 
    \begin{sc}
    \vspace{-.1cm}
    \resizebox{1.0\textwidth}{!}{\begin{tabular}{lccc}
    \toprule
     Method & Masking Distribution & Weighting Function & Conditional Distribution \\
     & $q(\vx_t|\vx_0)$ & $w(t)$ & $p_{\boldsymbol{\theta}}(\vx_0^i \mid \vx_t)$ \\
    \midrule
    MaskGIT & Uniformly mask $\lceil N \gamma_t \rceil$ tokens w/o replacement & $w(t) = 1$ & Categorical Distribution \\
    \midrule
    MAR &  Uniformly mask $\lceil N \gamma_t \rceil$ tokens w/o replacement & $w(t) = 1$ & Diffusion Model \\
    \midrule
    MDM & Mask $N$ tokens independently with ratio $\gamma_t$ & $w(t) = \frac{\gamma_t^\prime}{\gamma_t}$ & Categorical Distribution \\
    \bottomrule
    \end{tabular}}
    \end{sc} 
    \end{center}
    \vskip -0.1in
    \label{tab:unified_modeling}
\end{table*}

During sampling, MaskGIT starts with an image where all tokens are masked, $ \boldsymbol{x}_0 $. For each iteration $ t \in \{1, 2, \dots, T\} $, the number of masked tokens is $ n_{t} = \lceil \gamma_\frac{t}{T} N \rceil $, and the model receives input $ \boldsymbol{x}_{\frac{t-1}{T}} $. The model predicts the probabilities for all tokens, and the $ \hat{n}_t = n_{t-1} - n_t $ tokens with the highest confidence are unmasked, updating to $ \boldsymbol{x}_{\frac{t}{T}} $.

\textbf{MAR}~\cite{li2024autoregressive} proposes using a diffusion model~\cite{sohl2015deep} to model the per-token distribution, which eliminates the need for discrete tokenizers. By avoiding the information loss of discrete tokenizers, MAR achieves excellent image generation performance. 

During training, MAR samples the masking ratio $m_r$ from a truncated Gaussian distribution with mean 1.0, standard deviation 0.25, truncated to [0.7, 1.0].
For sampling, MAR adopts a decoding strategy similar to that of MaskGIT.

\subsection{Masked Diffusion Models}
Let $ \boldsymbol{x} = [\boldsymbol{x}^i]_{i=1}^N $ represent the discrete text tokens of a sentence, $ \text{[M]} $ denote the special mask token, and $ \gamma_t $ represent the mask schedule. MDMs~\cite{loudiscrete, shi2024simplified, sahoo2024simple} gradually add masks to the data in the forward process and remove them during the reverse process. Here, we focus on the parameterized form of RADD~\cite{ou2024your}. Given a noise level $ t \in [0, 1] $, the forward process of MDM is defined as adding noise independently in each dimension:
\begin{equation}
q_{t|0}(\boldsymbol{x}_t|\boldsymbol{x}_0) = \prod_{i=0}^{N-1} q_{t|0}(\boldsymbol{x}_t^i|\boldsymbol{x}_0^i),
\end{equation}
where
\begin{equation}
q_{t|0}(\boldsymbol{x}_t^i|\boldsymbol{x}_0^i) = 
\begin{cases} 
1 - \gamma_t, & \boldsymbol{x}_t^i = \boldsymbol{x}_0^i, \\ 
\gamma_t, & \boldsymbol{x}_t^i = \text{[M]}.
\end{cases}
\end{equation}

The training objective of MDM is to optimize the upper bound of the negative log-likelihood of the masked tokens, which defined as:

\begin{equation}
    \mathcal{L}(\boldsymbol{x_0}) = \int_0^1 \frac{\gamma_t^\prime}{\gamma_t} \mathbb{E}_{q(\boldsymbol{x}_t|\boldsymbol{x}_0)} \left[ \sum_{\{i|\boldsymbol{x}_t^i=\text{[M]}\}} \skipSpace -\log p_{\boldsymbol{\theta}}(\boldsymbol{x}_0^i|\boldsymbol{x}_t) \right] dt.
\label{eq:mdm_loss}
\end{equation}

Interestingly, the explicit time input of MDM is theoretically redundant
\footnote{Unlike continuous state diffusion 
which require both $\vx_t$ and $t$ as inputs to the model input to denoise, the mask discrete diffusion operates by
using $p_{\boldsymbol{\theta}}(\boldsymbol{x}_0^i|\boldsymbol{x}_t)$
instead of $p_{\boldsymbol{\theta}}(\boldsymbol{x}_0^i|\boldsymbol{x}_t, t)$.
That's because the timestep dependence can be extracted as a weight coefficient outside of the cross-entropy loss.
} \cite{ou2024your},
and has also been empirically validated in image generation \cite{hu2024maskneed}.

During sampling, given two noise levels $ s $ and $ t $, where $ 0 \leq s < t \leq 1 $, the reverse process is characterized as:
\begin{equation}
q_{s|t}(\boldsymbol{x}_s|\boldsymbol{x}_t) = \prod_{i=0}^{N-1} q_{s|t}(\boldsymbol{x}_s^i|\boldsymbol{x}_t),
\end{equation}
where
\begin{equation}
q_{s|t}(\boldsymbol{x}_s^i|\boldsymbol{x}_t) =
\begin{cases} 
1, & \boldsymbol{x}_s^i = \boldsymbol{x}_t^i, \, \boldsymbol{x}_t^i \neq \text{[M]}, \\
\frac{\gamma_s}{\gamma_t}, & \boldsymbol{x}_s^i = \text{[M]}, \, \boldsymbol{x}_t^i = \text{[M]}, \\
\frac{\gamma_t - \gamma_s}{\gamma_t} q_{0|t}(\boldsymbol{x}_s^i|\boldsymbol{x}_t), & \boldsymbol{x}_s^i \neq \text{[M]}, \, \boldsymbol{x}_t^i = \text{[M]}, \\
0, & \text{otherwise.}
\end{cases}
\label{eq:mdm_sample}
\end{equation}

\section{Unifying Masked Image Generation}


After removing the explicit time input from MDM, we observe that the MaskGIT objective (Eq.~\ref{eq:maskgit_loss}) can be expressed in terms of the general MDM loss formulation (Eq.~\ref{eq:mdm_loss}). Specifically, the Monte Carlo expectation over \( r \) in Eq.~\ref{eq:maskgit_loss} is equivalent to integrating over \( t \) from 0 to 1, where \( r \) can be interpreted as a scaled time variable \( t \) corresponding to the masking schedule.  In this reinterpretation, the masked image \( \boldsymbol{x}_{\overline{\textbf{M}}} \) in MaskGIT can be understood as \( \vx_t \) in the general framework, representing the noisy or partially masked image at time \( t \). That is, the masking distribution \( q_{\gamma_r}(\boldsymbol{x}_{\overline{\textbf{M}}}|\boldsymbol{x}) \) can be mapped to a specific instance of \( q(\vx_t|\vx_0) \), characterized by the chosen mask scheduling function \( \gamma_t \). \emph{See the equivalence between these two masking distributions in Appendix~\ref{app:equival_maskgit_mdm}.} After aligning these two masking distributions,  MaskGIT, MAR, and MDM can be expressed within a unified loss function, defined as:
\begin{equation}
    \mathcal{L}(\boldsymbol{x}_0)\! =\! \! \int_{\tmin}^{\tmax}\! \! \! w(t) \mathbb{E}_{q(\vx_t|\vx_0)}\left[ \sum_{\{i|\vx_t^i=\text{[M]}\}}\skipSpace \!\!\! -\log p_{\boldsymbol{\theta}}\left( \vx_0^i \, | \, \vx_t \right)  \right]\!dt.
\end{equation}

In this unified formulation, the key differences between the models primarily lie in the three components outlined in~\cref{tab:unified_modeling}. We explain these components as follows:

\textbf{Masking distribution $q(\vx_t|\vx_0)$.} For MaskGIT and MAR, $\lceil N\gamma_t \rceil $ tokens are uniformly masked without replacement as $\text{[M]}$. For MDM, each of the $N$ tokens is masked with probability $\gamma_t$ independently.\\
\textbf{Weighting function $w(t)$.} The weight function $w(t)$ determines the importance of the loss at each time step. For MaskGIT and MAR, $w(t) = 1$; for MDM, $w(t) = \frac{\gamma_t^\prime}{\gamma_t}$.\\
\textbf{Conditional distribution $p_{\boldsymbol{\theta}}\left( \vx_0^i \, | \, \vx_t \right)$.} 
For MaskGIT and MDM, the conditional distribution \( p_{\boldsymbol{\theta}}\left( \vx_0^i \, | \, \vx_t \right) \) is modeled as a categorical distribution. 
In contrast, for MAR, we employ a diffusion model assisted by a latent variable \(\vz\), leading to the following formulation:
\begin{equation}
    p_{\boldsymbol{\theta}}(x_0^i|\vx_t) = \int \delta_{\boldsymbol{\theta_1}}(\vz^i|\vx_t) p^{\text{diff}}_{\boldsymbol{\theta_2}}(x_0^i|\vz^i) d\vz^i.
\end{equation}
Here, \( \delta_{\boldsymbol{\theta_1}}(\vz^i|\vx_t) \) represents the output of the mask prediction model with input \( \vx_t \), and \( p^{\text{diff}}_{\boldsymbol{\theta_2}}(x_0^i|\vz^i) \) donated the output of diffusion model conditioned on \(\vz^i\). 

\section{Investigating the Design Space of Training}
\begin{figure*}[t!]
    \centering
    \subfigure[Choices of mask schedule\label{fig:fid_mask_schedule}]{
        \includegraphics[width=0.3\linewidth]{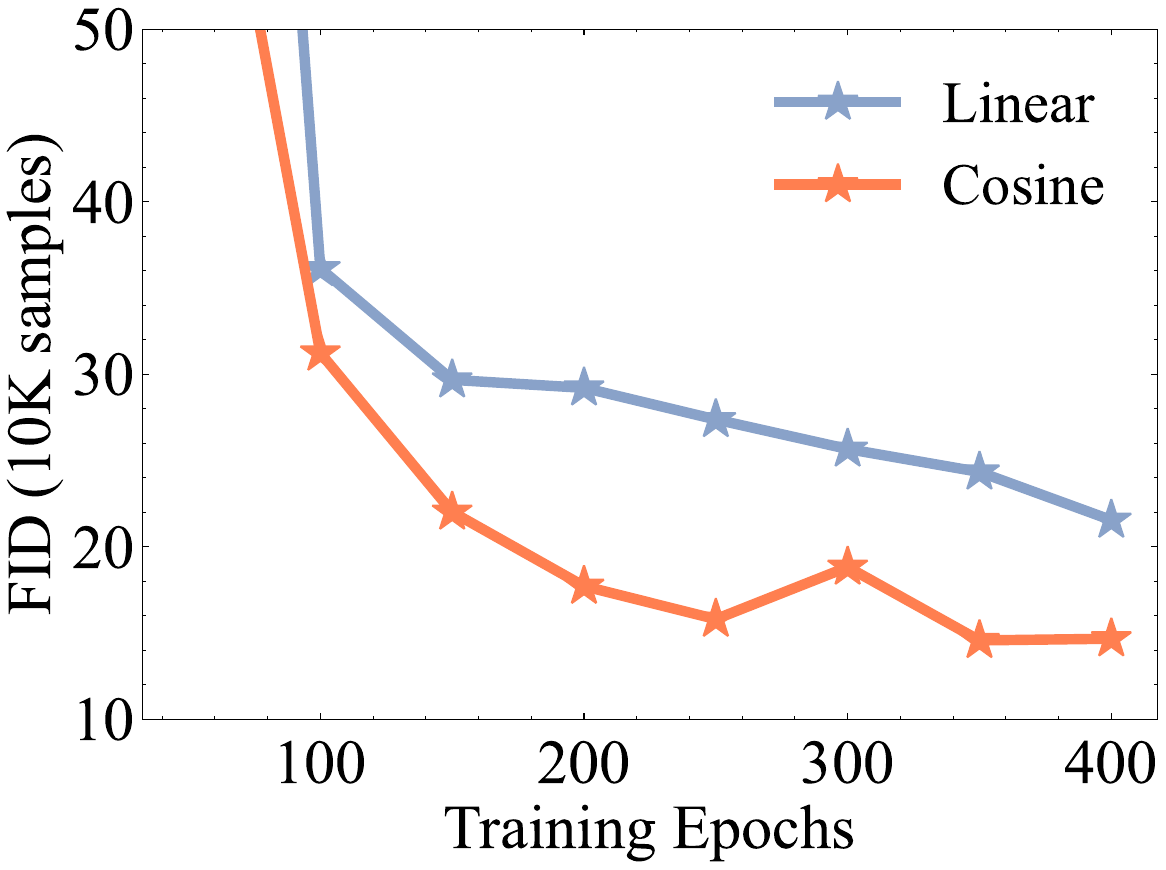}
    }
    \hspace{.1cm}
    \subfigure[Choices of weighting function\label{fig:fid_loss_weight}]{
        \includegraphics[width=0.3\linewidth]{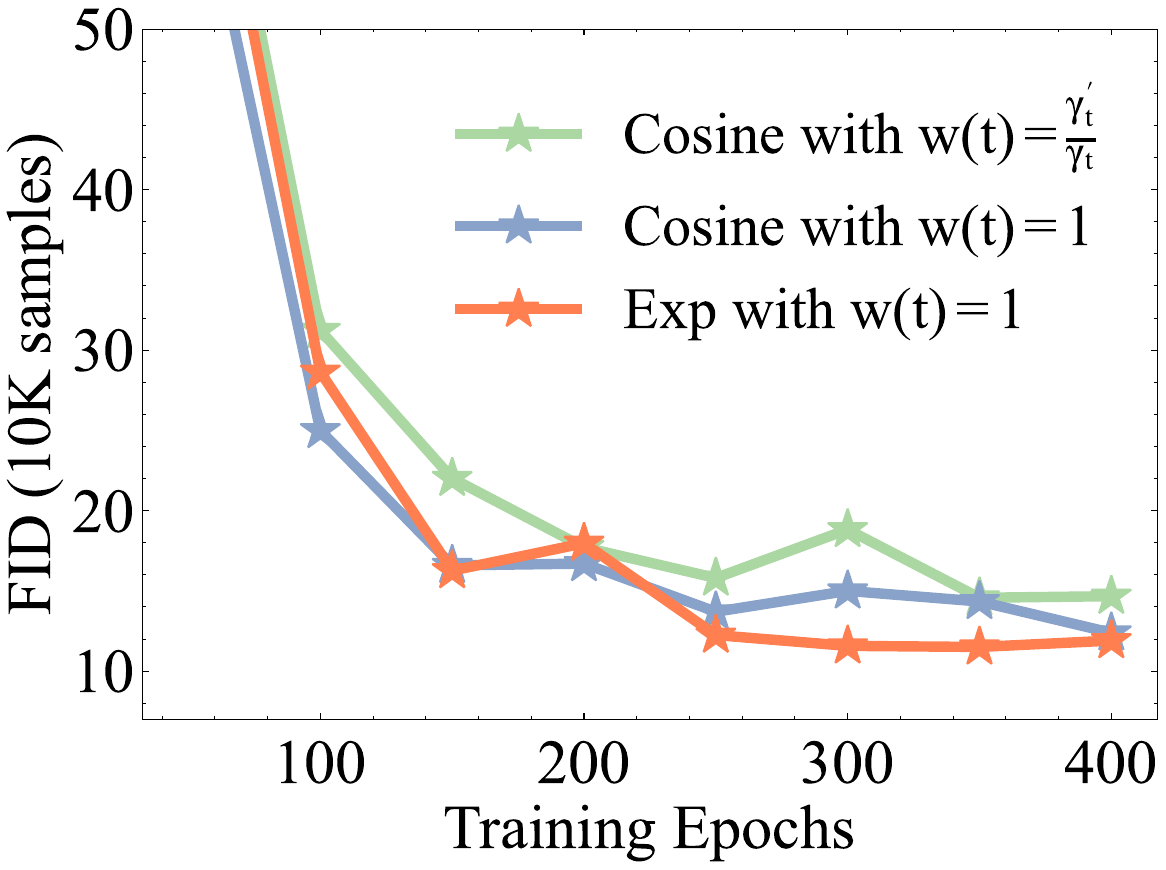}
    }
    \hspace{.1cm}
    \subfigure[Use the MAE trick\label{fig:fid_mae}]{
        \includegraphics[width=0.3\linewidth]{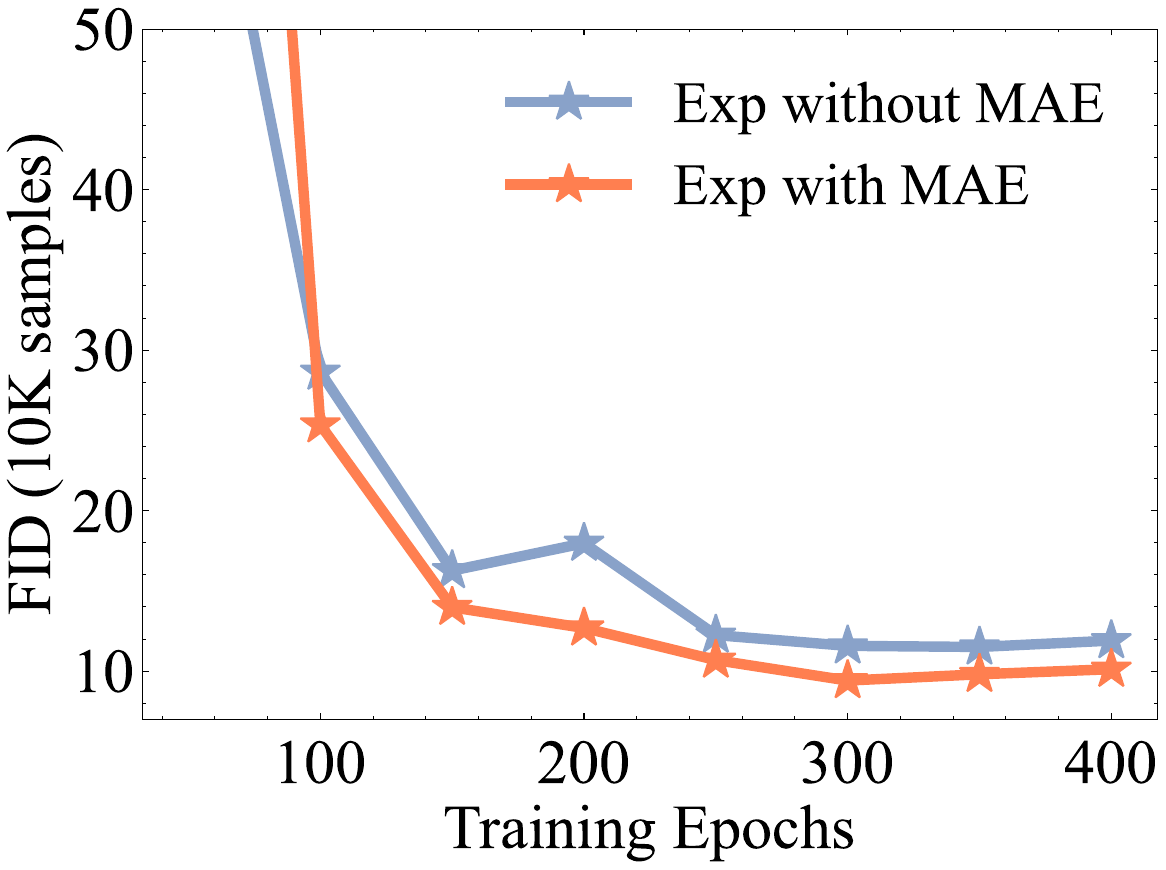}
    }\\
    \vspace{.2cm}
    \subfigure[Use the time truncation\label{fig:fid_t_trunc}]{
        \includegraphics[width=0.3\linewidth]{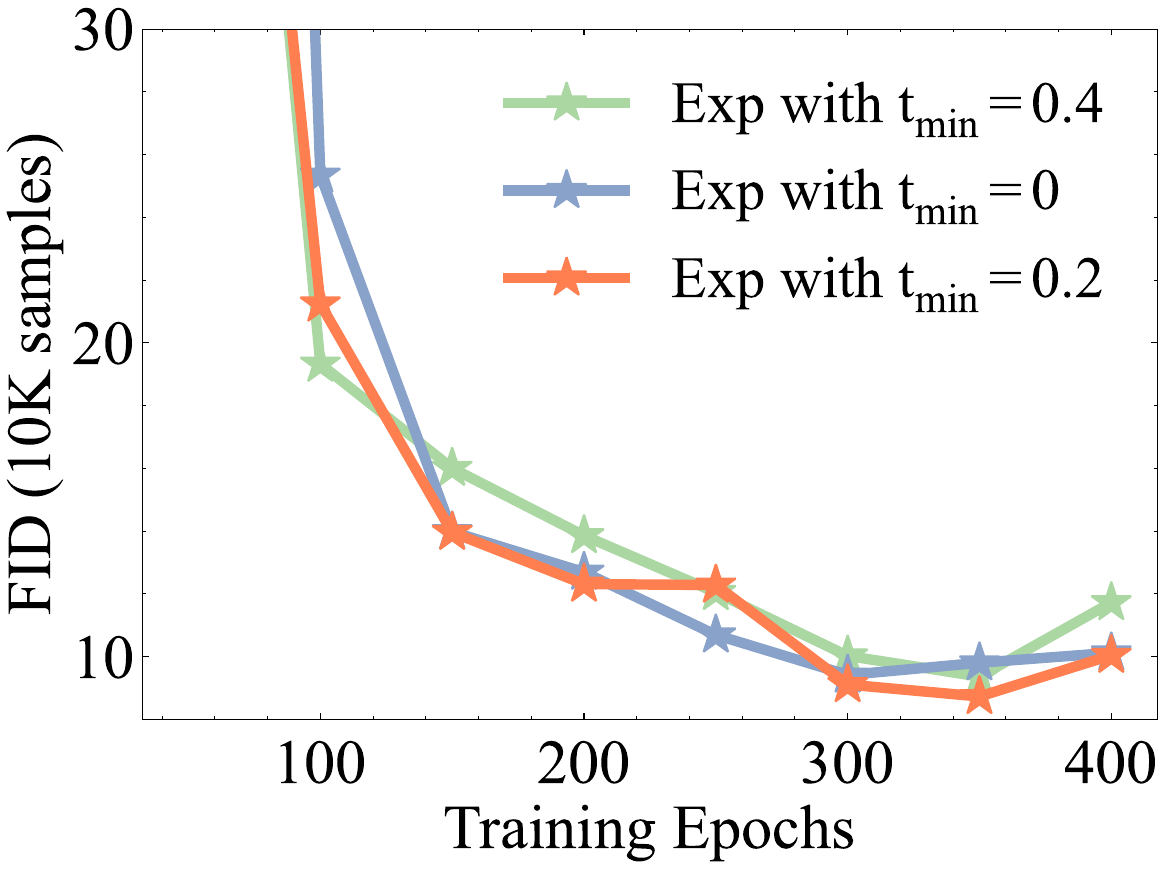}
    }
    \hspace{.1cm}
    \subfigure[Use CFG with mask\label{fig:fid_cfg_mask}]{
        \includegraphics[width=0.3\linewidth]{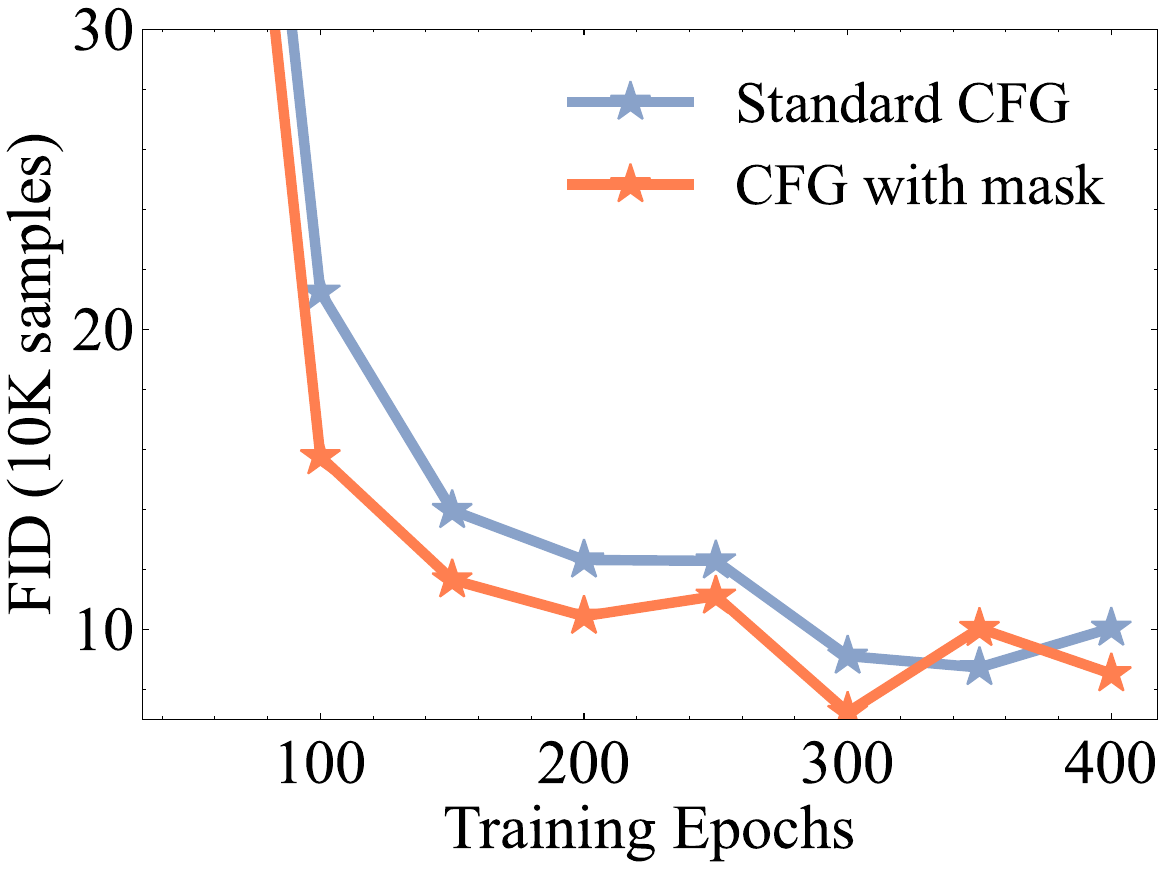}
    }
    \caption{\textbf{Exploring the design space of training.} Orange solid lines indicate the preferred choices in each subfigure.}
    \label{fig:ab}
\end{figure*}

Building upon the unified framework, we now explore various design choices within this formulation. Given the equivalence of masking distributions, we adopt MDM's as the default setting. Furthermore, to mitigate the information loss introduced by the discrete tokenizer~\cite{van2017neural, esser2021taming}, we use a diffusion model to model the conditional distribution \( p_{\boldsymbol{\theta}}(x_0^i|\vx_t) \). Our exploration begins with the standard MDM, which utilizes a single encoder transformer architecture and a linear mask schedule, in addition to using the diffusion model to model the conditional distribution \( p_{\boldsymbol{\theta}}\left( \vx_0^i \, | \, \vx_t \right) \). 

\textbf{Mask schedule.} The first critical aspect of our exploration is the choice of $\gamma_t$, which determines the probability of masking each token during the forward process (See Appendix~\ref{app:mask_schedule} for details). In this section, we use the weighting function of $w(t) = \frac{\gamma_t^\prime}{\gamma_t}$, which is mainly used in MDM. We consider three mask schedules: (1) \emph{Linear}: $\gamma_t = t$; (2) \emph{Cosine}: $\gamma_t = \cos\left(\frac{\pi}{2}(1-t)\right)$; (3) \emph{Exp}: $\gamma_t = 1 - \exp(-5t)$. The first two mask schedules are also mentioned in \citet{shi2024simplified}, while the last one is our design to achieve a higher masking ratio during training. As shown in Fig.~\ref{fig:fid_mask_schedule}, the cosine schedule outperforms the linear schedule. We hypothesize that, due to the high information redundancy in images, the cosine schedule achieves a higher mask ratio during training, providing stronger learning signals and leading to improved performance. The exp schedule further increases the mask ratio but destabilizes MDM training, likely due to the persistently large weighting function \( w(t) \), even at high mask ratios (see Fig.~\ref{fig:mask_schedule} for visualization of $w(t)$ and $\gamma_t$).

\textbf{Weighting function.} We consider two choices for $w(t)$. (1) $w(t) = \frac{\gamma_t^\prime}{\gamma_t}$, as used in MDM; (2) $w(t) = 1$, as used in MaskGIT. Notably, the weighting function significantly affects the choice of mask schedule. For instance, using $w(t) = \frac{\gamma_t^\prime}{\gamma_t}$ led to unstable training, particularly with the exp schedule. In contrast, as shown in Fig.~\ref{fig:fid_loss_weight}, setting $w(t)=1$ stabilized the training process and improved performance, similar to the phenomenon observed in DDPM~\cite{ho2020denoising}; under this setting, the exp schedule yielded the best results. Therefore, we adopted this combination ($w(t)=1$ and the exp schedule) as our default.

\textbf{Model Architecture.} We consider two model architectures: (1) A single-encoder transformer; (2) The MAE~\cite{he2022masked} architecture, which decomposes the transformer into an encoder-decoder structure, where the encoder processes only unmasked tokens. The primary difference between these architectures is whether the encoder receives masked tokens as input. As shown in Fig.~\ref{fig:fid_mae}, under the exp schedule, the MAE architecture outperforms the single-encoder transformer. Interestingly, despite being originally designed for self-supervised learning, MAE retains its advantages in image generation. Therefore, unless otherwise specified, we adopt the MAE architecture as the default setting.

\textbf{Time Truncation.} To achieve a higher mask ratio during training, in addition to selecting a more concave function for $\gamma_t$, we can also use time truncation, which restricts the minimum value of \( t \) to \( t_{\text{min}} \). We consider three choices: (1) $t_{\text{min}} = 0$, the original design; (2) $t_{\text{min}} = 0.2$; (3) $t_{\text{min}} = 0.4$. As shown in Fig.~\ref{fig:fid_t_trunc}, we observed that an appropriate time truncation ($t_{\text{min}} = 0.2$) can be beneficial and accelerates training convergence. However, excessive truncation (\( t_{\text{min}} = 0.4 \), where over 80\% of image tokens are masked during training) provides no benefit and may even degrade performance compared to no time truncation. Unless otherwise noted, we adopt \( t_{\text{min}} = 0.2 \) as the default setting.

\textbf{CFG with Mask.} Classifier-Free Guidance (CFG)~\cite{ho2022classifier} is widely used for guiding continuous diffusion models and masked image generation. It combines outputs of a conditional model (with class information) and an unconditional model (without class information) to improve alignment with the conditional output. In standard CFG, the unconditional model typically receives a learnable fake class token as input. Unsupervised classifier-free guidance was initially developed for text generation~\cite{nie2024scaling}, a process involving the unconditional model receiving a special mask token as input. Inspired by this method, our paper adapts it for image generation. We term this adapted approach \emph{CFG with Mask} to emphasize its focus on masked image generation. As shown in Fig.~\ref{fig:fid_cfg_mask}, CFG with mask improves generation performance compared to standard CFG. Notably, here we use only simple conditional generation without guidance, our results suggest that using a fake class token negatively impacts the conditional generation performance of MDM. Thus, we adopt CFG with mask as the default setting.

\section{Investigating the Design Space of Sampling}
\begin{figure*}[t!]
    \centering
    \subfigure[Choices of sample mask schedule\label{fig:sample_fid_mask_schedule}]{
        \includegraphics[width=0.312\linewidth]{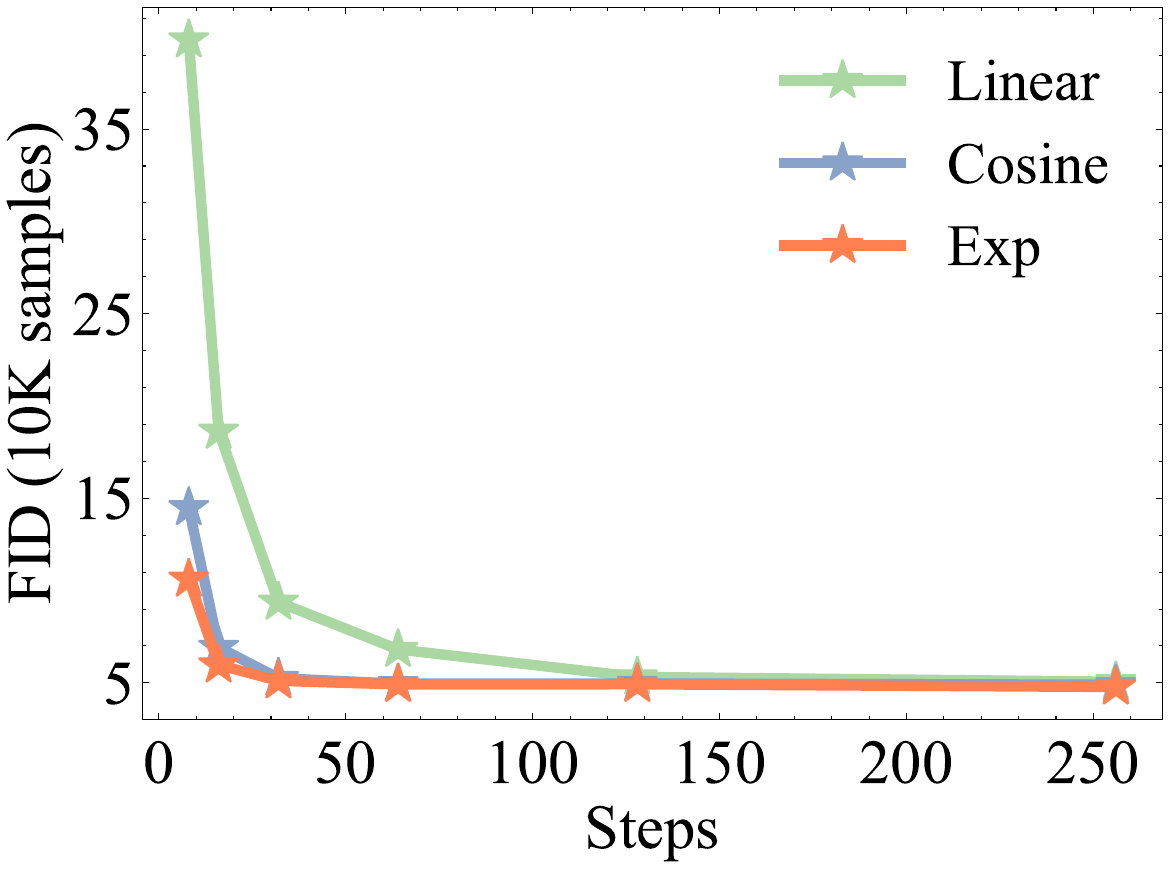}
    }
    \hspace{.05cm}
    \subfigure[Use the DPM-Solver\label{fig:sample_fid_dpm_solver}]{
        \includegraphics[width=0.312\linewidth]{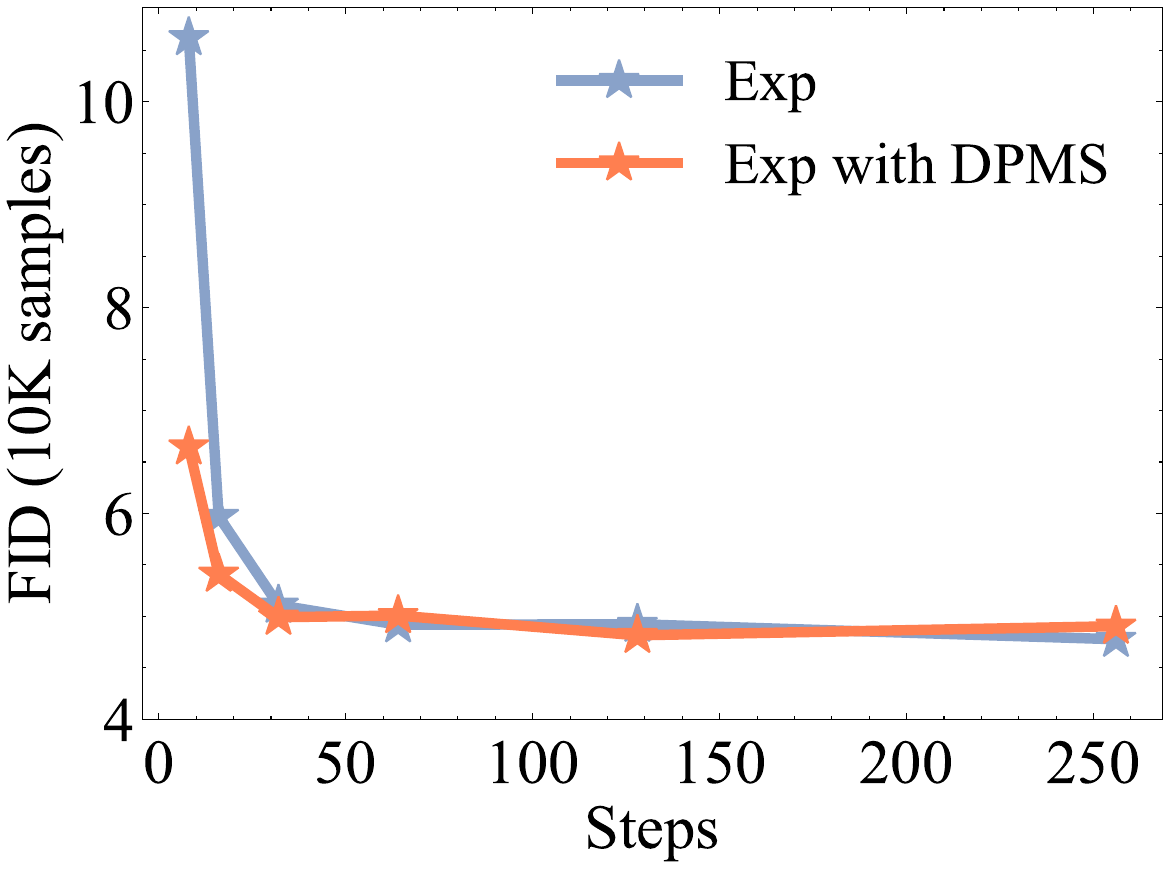}
    }
    \hspace{.05cm}
    \subfigure[Use the time interval\label{fig:sample_design}]{
        \includegraphics[width=0.312\linewidth]{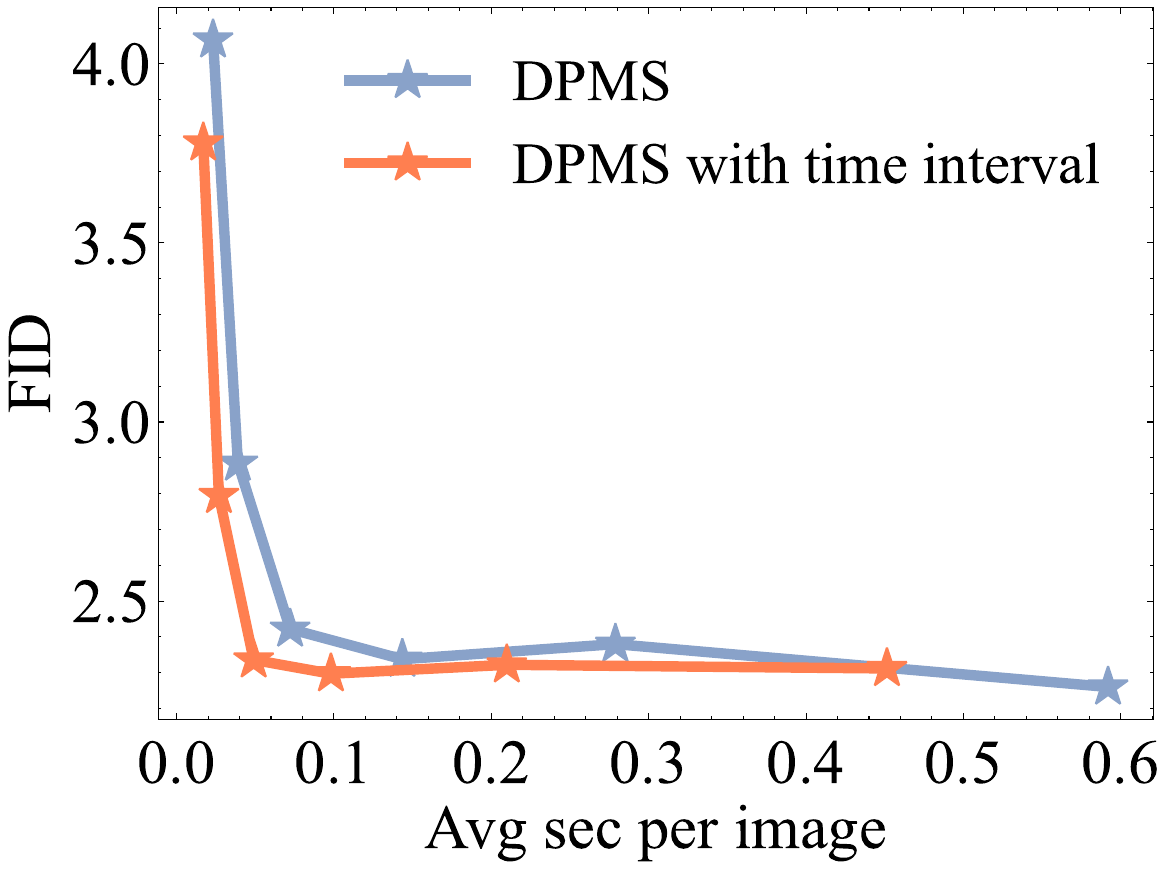}
    }
    \caption{\textbf{Exploring the design space of sampling.}  For each plot, points from left to right correspond to an increasing number of mask prediction steps: 8, 16, 32, and up to 256. In each subfigure, DPM-Solver is donated as DPMS. (a) The exp schedule outperforms others by predicting fewer tokens early. (b) DPM-Solver performs better with fewer prediction steps. (c) The time interval maintains performance while reducing sampling cost for each mask prediction step, particularly for high mask prediction steps.}
    \label{fig:sample_ab}
\end{figure*}
In the previous section, we carefully explore the training design space. In the following sections, we investigate the sampling design space. On one hand, we expect the model's performance to improve as the number of mask prediction steps increases. On the other hand, we aim to maintain strong performance even with a low number of mask prediction steps (e.g., 16).

\subsection{Mask Schedule during Sampling} 
During training, we observe that the exp schedule achieves the best performance. However, during sampling, different schedules may be employed. We are interested in identifying which mask schedule can achieve both of our goals.

To this end, we first conduct a simulation experiment (see details in Appendix~\ref{app:sample_similator_exp}) to compare the number of tokens predicted during each mask prediction step across different mask schedules. We observe that the linear schedule predicts a nearly constant number of tokens per step, while the cosine schedule predicts fewer tokens early in the process and progressively more later. This observation aligns with the findings reported in \citet{shi2024simplified}. Besides, the exp schedule predicts even fewer tokens initially, with a more gradual increase as the process continues.  As shown in Fig.~\ref{fig:sample_fid_mask_schedule}, we observe that each mask schedule benefits more prediction steps. Moreover, for low mask prediction steps (e.g., 8 or 16), the exp schedule outperforms the cosine schedule, which in turn outperforms the linear schedule. This suggests that, in the early stages of sampling, predicting fewer tokens may contribute to improved performance at lower mask prediction steps. Thus, we adopt the exp schedule as our default for sampling unless otherwise specified.

\subsection{The Sampling Method of Diffusion Loss} 
We use the diffusion loss to model the distribution of $p_{\boldsymbol{\theta}}\left( \vx_0^i \, | \, \vx_t \right)$. Previously, we follow MAR~\cite{li2024autoregressive} and use DDPM~\cite{ho2020denoising} sampling method with 100 diffusion steps. Additionally, MAR employs the temperature $\tau$ sampling method from ADM~\cite{dhariwal2021diffusion} to scale the noise by $\tau$, which requires careful tuning for optimal performance.

In contrast, DPM-Solver~\cite{lu2022dpm, lu2022dpm++} is a training-free, fast ODE sampler that accelerates the diffusion sampling process and converges faster with fewer steps. Interestingly, although DPM-Solver is designed for accelerating the diffusion process, we observe that, with low mask prediction steps, it outperforms DDPM, as shown in Fig.~\ref{fig:sample_fid_dpm_solver}. For example, with 8 mask prediction steps, DPM-Solver achieves an FID of 6.6, while DDPM, with a temperature of 1.0, achieves an FID of 10.6. We hypothesize that for low mask prediction steps, DDPM requires careful temperature tuning, whereas DPM-Solver, being an ODE sampler, does not require such adjustments. Moreover, DPM-Solver achieves good performance with fewer than 15 diffusion steps, while DDPM requires 100 diffusion steps. Therefore, unless specified, we default to DPM-Solver.

\begin{figure*}[t]
    \centering
    \subfigure[FLOPs vs. FID across model scales.\label{fig:scaling_flops_fid_scatter}]{
        \includegraphics[width=0.312\linewidth]{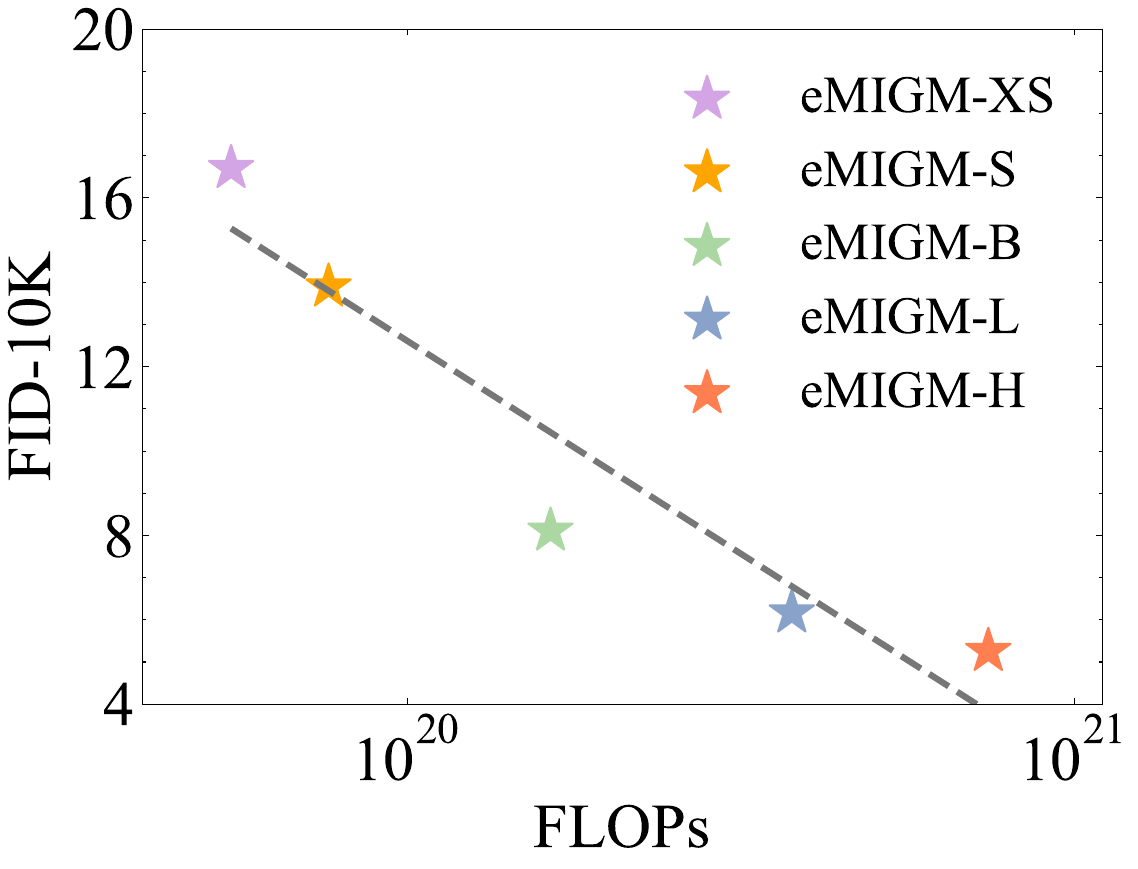}
    }
    \hspace{.05cm}
    \subfigure[FLOPs vs. FID under different budgets.\label{fig:scaling_flops_fid}]{
        \includegraphics[width=0.312\linewidth]{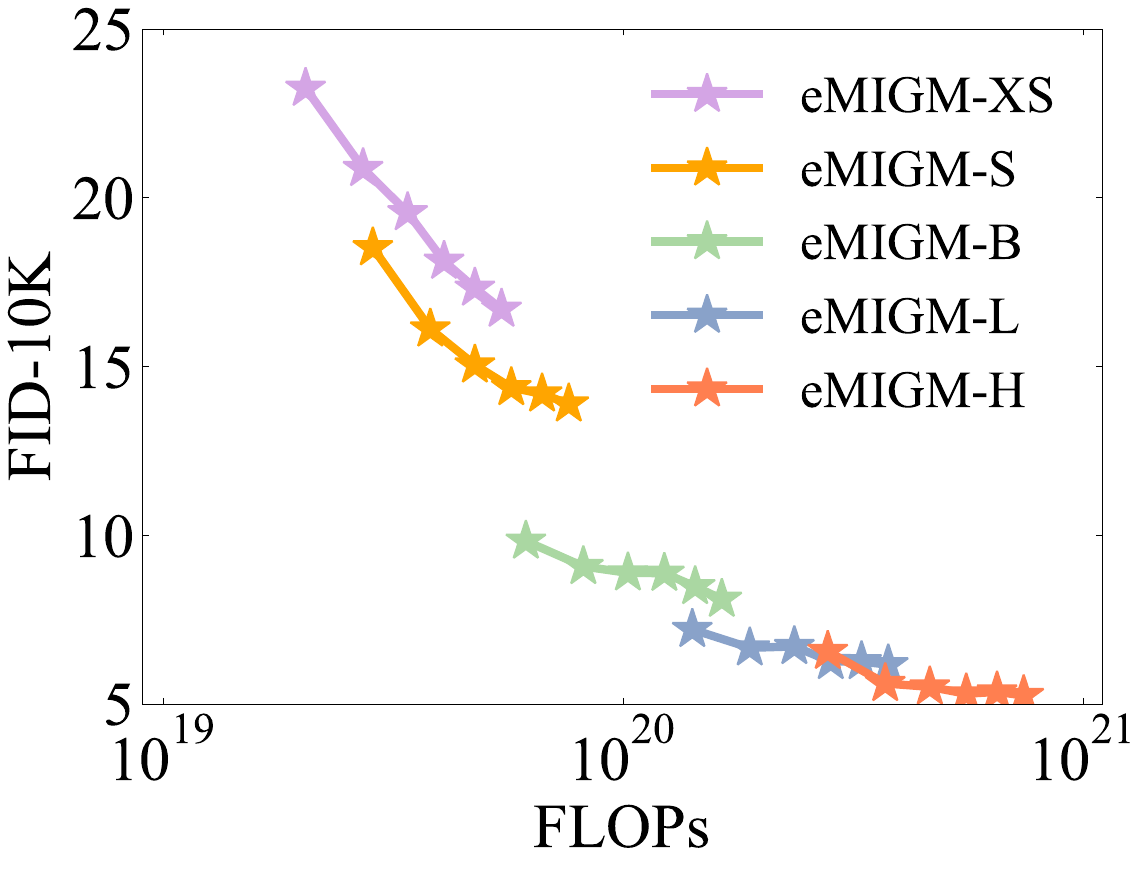}
    }
    \hspace{.05cm}
    \subfigure[Inference speed vs. FID.\label{fig:sample_scaling}]{
        \includegraphics[width=0.312\linewidth]{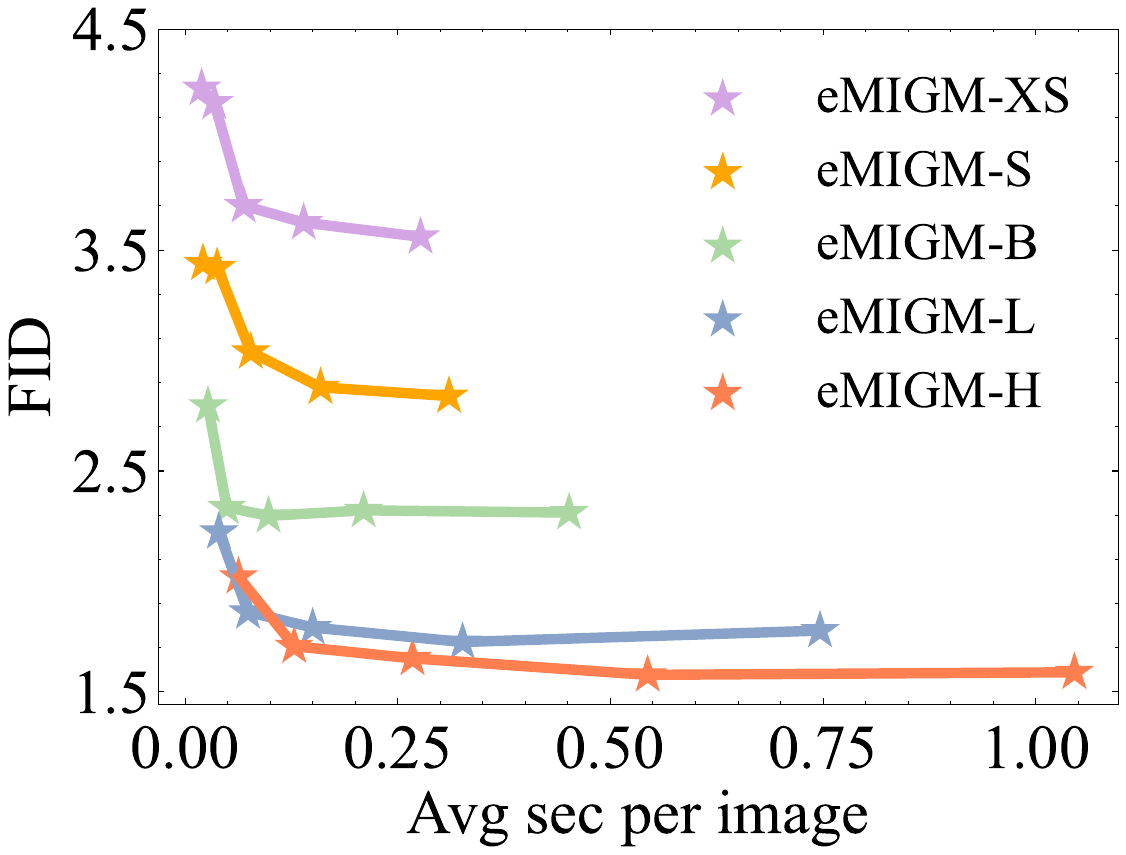}
    }
    \caption{\textbf{Scalability of eMIGM.} (a) A negative correlation demonstrates that eMIGM benefits from scaling. (b) Larger models are more training-efficient (i.e., achieving better sample quality with the same training FLOPs). (c) Larger models are more sampling-efficient (i.e., achieving better sample quality with the same inference time).}
    \label{fig:scaling}
\end{figure*}

\begin{table*}[!t]
\centering
\caption{\textbf{Image generation results on ImageNet $256\times256$.} $^\dagger$ denotes results taken from MaskGIT~\cite{chang2022maskgit}, and $^\star$ indicates results that require assistance from the self-supervised model. \emph{With $42\%$ of function evaluations (NFE), eMIGM-H achieves performance comparable to the state-of-the-art diffusion model REPA~\cite{yu2024representation}.} We \textbf{bold} the best result under each method and \underline{underline} the second-best result.}
\vspace{5pt} 
\label{tab:in256}
\begin{minipage}{0.48\textwidth}
\resizebox{\textwidth}{!}{%
\begin{tabular}{lccc}
\toprule
\textbf{METHOD} & \textbf{NFE} ($\downarrow$) & \textbf{FID} ($\downarrow$) & \textbf{\#Params} \\
\midrule
\multicolumn{4}{l}{\textbf{Diffusion models}} \\
\midrule
ADM-G~\cite{dhariwal2021diffusion}& 250$\times$ 2 & 4.59 & 554M\\
ADM-G-U~\cite{dhariwal2021diffusion}& 750 & 3.94 & 554M\\
LDM-4-G~\cite{rombach2022high}& 250$\times$ 2 & 3.60 & 400M\\
VDM++~\cite{kingma2024understanding}&512$\times$2 & 2.40 & 2B \\
SimDiff~\cite{hoogeboom2023simple}& 512$\times$2 & 2.44 &2B \\
U-ViT-H/2~\cite{bao2023all}& 50$\times$2 &2.29& 501M\\
DiT-XL/2~\cite{peebles2023scalable} & 250$\times$2 & 2.27 & 675M\\
Large-DiT~\cite{alpha2024large} &  250$\times$2 & 2.10 & 3B \\
Large-DiT~\cite{alpha2024large} &  250$\times$2 & 2.28 & 7B \\
SiT-XL~\cite{ma2024sit} & 250$\times$2 & 2.06 & 675M\\
D$_\text{IFFU}$SSM-XL-G~\cite{yan2024diffusion} & 250$\times$2 & 2.28 & 660M \\
DiffiT~\cite{hatamizadeh2025diffit} &250$\times$2 & \underline{1.73} & 561M \\
REPA~\cite{yu2024representation}$^\star$ & 250$\times$1.7 & \textbf{1.42} & 675M \\
\midrule
\multicolumn{4}{l}{\textbf{ARs}} \\
\midrule
VQGAN~\cite{esser2021taming}$^\dagger$ & 256 & 18.65 & 227M \\
VAR-$d16$~\cite{tian2024visual} & 10$\times$2 & 3.30 & 310M\\
VAR-$d20$~\cite{tian2024visual} & 10$\times$2 & 2.57 & 600M\\
VAR-$d24$~\cite{tian2024visual} & 10$\times$2 & \underline{2.09} & 1B\\
VAR-$d30$~\cite{tian2024visual} & 10$\times$2 & \textbf{1.92} & 2B\\
\bottomrule
\end{tabular}
}
\end{minipage}
\begin{minipage}{0.48\textwidth}
\resizebox{\textwidth}{!}{%
\begin{tabular}{lccc}
\toprule
\textbf{METHOD} & \textbf{NFE} ($\downarrow$) & \textbf{FID} ($\downarrow$) & \textbf{\#Params} \\
\midrule 
\multicolumn{4}{l}{\textbf{GANs}} \\
\midrule
BigGAN~\cite{brock2018large} & 1 & 6.95 & - \\
StyleGAN-XL~\cite{sauer2022stylegan} & 1$\times$2 & 2.30 & -  \\
\midrule
\multicolumn{4}{l}{\textbf{Masked models}} \\
\midrule
MaskGIT~\cite{chang2022maskgit}$^\dagger$~~~~~~~~~~~ & 8 & 6.18 & 227M \\ 
MAR-B~\cite{li2024autoregressive} & 256$\times2$ & 2.31 & 208M \\
MAR-L~\cite{li2024autoregressive} & 256$\times2$ & 1.78 & 479M \\
MAR-H~\cite{li2024autoregressive} & 256$\times2$ & \textbf{1.55} & 943M \\
\midrule
\multicolumn{4}{l}{\textbf{Ours}} \\
\midrule
eMIGM-XS & 16$\times$1.2 & 4.23 & 69M\\ 
eMIGM-S & 16$\times$1.2 & 3.44 & 97M\\ 
eMIGM-B & 16$\times$1.2 & 2.79 & 208M\\ 
eMIGM-L & 16$\times$1.2 & 2.22 & 478M\\
eMIGM-H & 16$\times$1.2 & 2.02 & 942M\\
\midrule
eMIGM-XS & 128$\times$1.4 & 3.62 & 69M\\ 
eMIGM-S & 128$\times$1.4 & 2.87 & 97M\\ 
eMIGM-B & 128$\times$1.35 & 2.32 & 208M\\ 
eMIGM-L & 128$\times$1.4 & 1.72 & 478M\\
eMIGM-H & 128$\times$1.4 & \underline{1.57} & 942M\\
\bottomrule
\end{tabular} %
}

\end{minipage}
\end{table*}

\begin{table*}[t]
\centering
\caption{\textbf{Image generation results on ImageNet $512\times512$.} $^\dagger$ denotes results taken from MaskGIT~\cite{chang2022maskgit}. 
$^\ddagger$ denotes results obtained using Guidance Interval~\cite{kynkaanniemi2024applying}. \emph{With 20 function evaluations (NFE), eMIGM-L outperforms strong visual autoregressive models VAR~\cite{tian2024visual}. When the NFE increases to 80, eMIGM-L surpasses the strong diffusion model EDM2~\cite{karras2024analyzing}.} We \textbf{bold} the best result under each method and \underline{underline} the second-best result.}
\vspace{5pt} 
\label{tab:in512}
\begin{minipage}{0.48\textwidth}
\resizebox{\textwidth}{!}{%
\begin{tabular}{lccc}
\toprule
\textbf{METHOD} & \textbf{NFE} ($\downarrow$) & \textbf{FID} ($\downarrow$) & \textbf{\#Params} \\
\midrule
\multicolumn{4}{l}{\textbf{Diffusion models}} \\
\midrule
ADM-G~\cite{dhariwal2021diffusion}& 250$\times$ 2 & 7.72 & 559M\\
ADM-G-U~\cite{dhariwal2021diffusion}& 750 & 3.85 & 559M\\
VDM++~\cite{kingma2024understanding}&512$\times$2 &2.65 &2B \\
SimDiff~\cite{hoogeboom2023simple}& 512$\times$2 &3.02 &2B \\
U-ViT-H/4~\cite{bao2023all}& 50$\times$2 &4.05& 501M\\
DiT-XL/2~\cite{peebles2023scalable} & 250$\times$2& 3.04 & 675M\\
Large-DiT~\cite{alpha2024large} &  250$\times$2 & 2.52 & 3B \\
SiT-XL~\cite{ma2024sit} & 250$\times$2 & 2.62 & 675M\\
EDM2-XXL~\cite{karras2024analyzing} & 32$\times$2 & \underline{1.81} & 1.5B\\
EDM2-XXL~\cite{kynkaanniemi2024applying}$^\ddagger$ & 32$\times$1.2 & \textbf{1.40} & 1.5B \\
\midrule
\multicolumn{4}{l}{\textbf{Consistency models}} \\
\midrule
sCT-XXL~\cite{lu2024simplifying} & 2 & 3.76 & 1.5B \\
sCD-XXL~\cite{lu2024simplifying} & 2 & 1.88 & 1.5B \\
\midrule 
\multicolumn{4}{l}{\textbf{GANs}} \\
\midrule
BigGAN~\cite{brock2018large} & 1 & 8.43 & - \\
StyleGAN-XL~\cite{sauer2022stylegan} & 1$\times$2 & 2.41 & -  \\

\bottomrule
\end{tabular}
}
\end{minipage}
\begin{minipage}{0.48\textwidth}
\resizebox{\textwidth}{!}{%
\begin{tabular}{lccc}
\toprule
\textbf{METHOD} & \textbf{NFE} ($\downarrow$) & \textbf{FID} ($\downarrow$) & \textbf{\#Params} \\
\midrule
\multicolumn{4}{l}{\textbf{ARs}} \\
\midrule
VQGAN~\cite{esser2021taming}$^\dagger$ & 1024 & 26.52 & 227M \\
VAR-$d36$-s~\cite{tian2024visual} & 10$\times$2 & 2.63 & 2.3B\\

\midrule
\multicolumn{4}{l}{\textbf{Masked models}} \\
\midrule
MaskGIT~\cite{chang2022maskgit}$^\dagger$~~~~~~~~ & 12 & 7.32 & 227M \\ 
MAR~\cite{li2024autoregressive} & 256$\times2$ & \textbf{1.73} & 481M \\

\midrule
\multicolumn{4}{l}{\textbf{Ours}} \\
\midrule
eMIGM-XS & 16$\times$1.2 & 4.63 & 104M\\ 
eMIGM-S & 16$\times$1.2 & 3.65 & 132M\\ 
eMIGM-B & 16$\times$1.2 & 2.78 & 244M\\ 
eMIGM-L & 16$\times$1.2 & 2.19 & 478M\\
\midrule
eMIGM-XS & 64$\times$1.25 & 4.45 &104M\\ 
eMIGM-S & 64$\times$1.25 & 3.29 & 132M\\ 
eMIGM-B & 64$\times$1.25 & 2.31 & 244M\\ 
eMIGM-L & 64$\times$1.25 & \underline{1.77} & 478M\\
\bottomrule
\end{tabular} %
}

\end{minipage}
\end{table*}
\subsection{Time Interval for Classifier Free Guidance} 
Previously, we adopt a linear CFG schedule following MAR~\cite{li2024autoregressive}, where the CFG value gradually increased from 0 to the target value during the mask prediction process. With a constant CFG schedule, we find that the generation performance is highly sensitive to the CFG value, as shown in Fig.~\ref{fig:cfg_ab}. We hypothesize that, for MDM, token generation is irreversible—once a token is generated, it cannot be modified. Therefore, a strong guide in the early stages may reduce the variation in the results, leading to a higher FID. This is similar to our earlier observation with the linear mask schedule, where generating too many incorrect tokens early can cause error accumulation and degrade the performance. We conduct an experiment with a total of 256 sample tokens and 16 mask prediction steps (see details in Appendix~\ref{app:time_interval_exp}) to validate our hypothesis. Let $s_i$ and $t_i$ denote the endpoint and start of the $i$-th step in the mask prediction process. We apply CFG if $s_i \in [\text{cfg}_{t_{\min}}, \text{cfg}_{t_{\max}}]$; otherwise, we use simple conditional generation. As shown in Fig.~\ref{fig:time_interval_fid}, when $\text{cfg}_{t_{\min}} < \text{cfg}_{t_{\max}} \leq 0.5$, we achieve a relatively low FID, supporting our hypothesis. In particular, the best performance is achieved when $\text{cfg}_{t_{\min}} = 0.1$ and $\text{cfg}_{t_{\max}} = 0.3$, using only 60\% of the NFE (the number of function evaluations) compared to standard CFG. Specifically, for standard CFG, NFE = $16 \times 2$, while for the time interval, NFE $\approx 16 + 16 \times (0.3 - 0.1)$. 

As shown in Fig.~\ref{fig:sample_design}, we observe that the time interval maintains performance at each mask prediction step while reducing sampling time. This demonstrates its efficiency and effectiveness. Therefore, we adopt the time interval for all subsequent experiments in this paper.

\section{Experiments}

By fully considering the design space mentioned above, we evaluate eMIGM on ImageNet $256\times256$ and ImageNet $512\times512$~\cite{deng2009imagenet}, benchmarking the sample quality using Fréchet Inception Distance (FID)~\cite{heusel2017gans}. See experiment settings in Appendix~\ref{app:exp_settings}.

\subsection{Larger Models Are Training and Sampling Efficient}
First, to demonstrate the scaling properties of eMIGM, we plot the FID-10K at 400 training epochs for different model sizes of eMIGM against training FLOPs. As shown in Fig.~\ref{fig:scaling_flops_fid_scatter}, we observe a negative correlation between training FLOPs and FID-10K, indicating that eMIGM benefits from scaling. Second, for different model sizes of eMIGM, we scale the FLOPs and analyze the FID-10K in relation to training FLOPs. As shown in Fig.~\ref{fig:scaling_flops_fid}, for each model size of eMIGM, as training epochs and training FLOPs increase, performance also improves. Additionally, we observe that for the same training FLOPs, larger eMIGM models achieve better performance. For instance, eMIGM-L outperforms eMIGM-B with approximately $10^{20}$ FLOPs. Third, we observed the inference-time scaling behavior of eMIGM. As shown in Fig.~\ref{fig:sample_scaling}, we plot the performance of different eMIGM model sizes across various mask prediction steps (ranging from 16 to 256). The speed is measured using a single A100 GPU with a batch size of 256. We observe that as the number of prediction steps increases, each model size of eMIGM achieves better performance, particularly for smaller models (i.e., eMIGM-XS and eMIGM-S). For larger model sizes, a similar best performance is reached with just 64 steps. Additionally, we also find that larger eMIGM models achieve better performance while maintaining similar inference speeds. For example, at a speed of about 0.2 seconds per image, eMIGM-L achieves a strong FID of 1.8, outperforming eMIGM-B with an FID of 2.3.

\subsection{Image Generation on ImageNet}
In Tab.~\ref{tab:in256}, we compare eMIGM with state-of-the-art generative models on \textbf{ImageNet $256\times256$}. Notably, in Tab.~\ref{tab:in256} and Tab.~\ref{tab:in512}, we list only the NFE of eMIGM's transformer component. When measured on a single A100 GPU with a batch size of 256, we found that the MLP diffusion block introduces approximately 14\% additional computational overhead beyond the NFE of the main transformer. However, since the transformer component remains the primary computational bottleneck, NFE continues to be a valid efficiency metric. By exploring the design space of sampling, eMIGM with few NFEs (approximately 20) outperforms VAR~\cite{tian2024visual} with a similar model size. Specifically, eMIGM-B achieves an FID of 2.79 with only 208M parameters, while VAR-d16 achieves an FID of 3.30 with 310M parameters. Notably, as we increase the NFE, all of our models consistently show significant improvements in generation performance. For instance, eMIGM-L achieves an FID of 1.72 with 180 NFEs, compared to an FID of 2.22 with 20 NFEs. By increasing the NFE, eMIGM-L, despite having only 478M parameters, outperforms the best VAR-d30, which achieves an FID of 1.92 with 2B parameters. Lastly, our more powerful eMIGM-H achieves an FID of 1.57 with just 180 NFEs, outperforming strong diffusion models such as Large-DiT~\cite{alpha2024large} and DiffiT~\cite{hatamizadeh2025diffit}. eMIGM-H is also comparable to the best diffusion models  REPA~\cite{yu2024representation}, which require 425 sequential steps and the assistance of the self-supervised model. Furthermore, compared to the state-of-the-art GAN model StyleGAN-XL~\cite{sauer2022stylegan}, eMIGM-B achieves superior performance. We also present more evaluation metrics on Tab.~\ref{tab:more_evaluaiton_metrics} in the appendix.

We also evaluate eMIGM on higher resolution images (i.e., $512 \times 512$) in Tab.~\ref{tab:in512}. Specifically, with similar NFEs, eMIGM-L (with only 478M parameters) achieves an FID of 2.19, outperforming the strong generative model VAR~\cite{tian2024visual} (with 2.3B parameters), which achieves an FID of 2.63. Furthermore, compared to the strong diffusion model EDM2~\cite{karras2024analyzing}, eMIGM-L achieves an FID of 1.77, outperforming EDM2's FID of 1.81. These quantitative results demonstrate that eMIGM achieves excellent generation performance and high sampling efficiency across diverse resolutions. However, when using the guidance interval~\cite{kynkaanniemi2024applying}, EDM2-XXL achieves superior performance while needing more parameters. A comparison of the sampling speeds for eMIGM and EDM2~\cite{karras2024analyzing} is also presented in Tab.~\ref{tab:sampling_speed_comparison}. Furthermore, when compared to MAR, eMIGM-L achieves competitive performance while using an NFE of less than 20\%.

\section{Related Work}
\textbf{Visual generation.} Modern visual generation models primarily fall into four categories: GANs~\cite{goodfellow2014generative, brock2018large, sauer2022stylegan}, diffusion models~\cite{song2020score, sohl2015deep, ho2020denoising}, masked prediction models~\cite{chang2022maskgit, li2023mage, li2024autoregressive, bai2024meissonic, shao2024bag, ni2024adanat}, and autoregressive models~\cite{esser2021taming, tian2024visual, sun2024autoregressive, tang2024hart}. The most related works to our study are MaskGIT~\cite{chang2022maskgit} and MAR~\cite{li2024autoregressive}. We provide a unified framework that integrates both approaches and systematically explore the impact of each component. Additionally, guidance interval~\cite{kynkaanniemi2024applying} and CADS~\cite{sadat2023cads} also observed that strong guidance early in the process negatively affects diversity. Therefore, they proposed sampling strategies to adjust the guidance application during sampling. Besides, \citet{wang2024analysis} also analyses the schedule of classifier-free guidance in continuous diffusion models. However, unlike our proposed time interval, which applies guidance at the token level, their methods operate at different noise levels of the entire image. Besides, our proposed time interval is motivated by MDM's unique irreversible token generation constraint. Furthermore, \citet{shao2024bag} proposed an enhanced inference technique to improve the speed and performance of masked image generative models such as MaskGIT~\cite{chang2022maskgit} and Meissonic~\cite{bai2024meissonic}. Their technique is orthogonal to our method and can also be applied to our work.

\textbf{Masked discrete diffusion models.}
Recently, masked discrete diffusion models~\cite{austin2023structured, campbell2022continuous}, a special case of discrete diffusion models~\cite{sohl2015deep, hoogeboom2021argmax}, have achieved remarkable progress in various domains, including text generation~\cite{he2022diffusionbert, lou2024discrete, shi2024simplified, sahoo2024simple, ou2024your, Zheng2023ARD, chen2023fast, gat2024discrete, nie2024scaling}, music generation~\cite{Sun2022ScorebasedCD}, protein design~\cite{campbell2024generativeflowsdiscretestatespaces}, and image generation~\cite{hu2024maskneed}.

\section{Conclusion}  
In this paper, we present a single framework to unify masked image generation models and masked diffusion models and carefully examine each component of design space to achieve efficient and high-quality image generation. Empirically, we demonstrate that eMIGM can achieve comparable performance with the state-of-the-art continuous diffusion models with fewer NFEs. We believe that eMIGM will inspire future research in masked image generation.

\section*{Acknowledgements}
This work was sponsered by the Beijing Nova Program (No. 20230484416); National Natural Science Foundation of China (No. 92470118); Beijing Natural Science Foundation (No. L247030); the ant Group Research Fund.

\section*{Impact Statement}
We introduce eMIGM, a powerful generative model that significantly accelerates the sampling speed while maintaining high image quality. However, this increased efficiency may increase the potential for misuse of generated images.  To mitigate this, watermarks can be embedded into the generated images without affecting the generation quality, helping to prevent misuse and verify if an image is generated.

\bibliography{icml2025}

@article{nie2024scaling,
  title={Scaling up Masked Diffusion Models on Text},
  author={Nie, Shen and Zhu, Fengqi and Du, Chao and Pang, Tianyu and Liu, Qian and Zeng, Guangtao and Lin, Min and Li, Chongxuan},
  journal={arXiv preprint arXiv:2410.18514},
  year={2024}
}

@article{ou2024your,
  title={Your Absorbing Discrete Diffusion Secretly Models the Conditional Distributions of Clean Data},
  author={Ou, Jingyang and Nie, Shen and Xue, Kaiwen and Zhu, Fengqi and Sun, Jiacheng and Li, Zhenguo and Li, Chongxuan},
  journal={arXiv preprint arXiv:2406.03736},
  year={2024}
}

@inproceedings{he2022masked,
  title={Masked autoencoders are scalable vision learners},
  author={He, Kaiming and Chen, Xinlei and Xie, Saining and Li, Yanghao and Doll{\'a}r, Piotr and Girshick, Ross},
  booktitle={Proceedings of the IEEE/CVF conference on computer vision and pattern recognition},
  pages={16000--16009},
  year={2022}
}

@article{ho2020denoising,
  title={Denoising diffusion probabilistic models},
  author={Ho, Jonathan and Jain, Ajay and Abbeel, Pieter},
  journal={Advances in neural information processing systems},
  volume={33},
  pages={6840--6851},
  year={2020}
}

@inproceedings{sohl2015deep,
  title={Deep unsupervised learning using nonequilibrium thermodynamics},
  author={Sohl-Dickstein, Jascha and Weiss, Eric and Maheswaranathan, Niru and Ganguli, Surya},
  booktitle={International conference on machine learning},
  pages={2256--2265},
  year={2015},
  organization={PMLR}
}

@inproceedings{loudiscrete,
  title={Discrete Diffusion Modeling by Estimating the Ratios of the Data Distribution},
  author={Lou, Aaron and Meng, Chenlin and Ermon, Stefano},
  booktitle={Forty-first International Conference on Machine Learning},
  year={2024}
}

@article{shi2024simplified,
  title={Simplified and Generalized Masked Diffusion for Discrete Data},
  author={Shi, Jiaxin and Han, Kehang and Wang, Zhe and Doucet, Arnaud and Titsias, Michalis K},
  journal={arXiv preprint arXiv:2406.04329},
  year={2024}
}

@article{sahoo2024simple,
  title={Simple and Effective Masked Diffusion Language Models},
  author={Sahoo, Subham Sekhar and Arriola, Marianne and Schiff, Yair and Gokaslan, Aaron and Marroquin, Edgar and Chiu, Justin T and Rush, Alexander and Kuleshov, Volodymyr},
  journal={arXiv preprint arXiv:2406.07524},
  year={2024}
}

@article{li2024autoregressive,
  title={Autoregressive Image Generation without Vector Quantization},
  author={Li, Tianhong and Tian, Yonglong and Li, He and Deng, Mingyang and He, Kaiming},
  journal={arXiv preprint arXiv:2406.11838},
  year={2024}
}

@article{van2017neural,
  title={Neural discrete representation learning},
  author={Van Den Oord, Aaron and Vinyals, Oriol and others},
  journal={Advances in neural information processing systems},
  volume={30},
  year={2017}
}

@inproceedings{esser2021taming,
  title={Taming transformers for high-resolution image synthesis},
  author={Esser, Patrick and Rombach, Robin and Ommer, Bjorn},
  booktitle={Proceedings of the IEEE/CVF conference on computer vision and pattern recognition},
  pages={12873--12883},
  year={2021}
}

@inproceedings{chang2022maskgit,
  title={Maskgit: Masked generative image transformer},
  author={Chang, Huiwen and Zhang, Han and Jiang, Lu and Liu, Ce and Freeman, William T},
  booktitle={Proceedings of the IEEE/CVF Conference on Computer Vision and Pattern Recognition},
  pages={11315--11325},
  year={2022}
}

@inproceedings{deng2009imagenet,
  title={Imagenet: A large-scale hierarchical image database},
  author={Deng, Jia and Dong, Wei and Socher, Richard and Li, Li-Jia and Li, Kai and Fei-Fei, Li},
  booktitle={2009 IEEE conference on computer vision and pattern recognition},
  pages={248--255},
  year={2009},
  organization={Ieee}
}

@article{devlin2018bert,
  title={Bert: Pre-training of deep bidirectional transformers for language understanding},
  author={Devlin, Jacob},
  journal={arXiv preprint arXiv:1810.04805},
  year={2018}
}

@article{ho2022classifier,
  title={Classifier-free diffusion guidance},
  author={Ho, Jonathan and Salimans, Tim},
  journal={arXiv preprint arXiv:2207.12598},
  year={2022}
}

@article{dhariwal2021diffusion,
  title={Diffusion models beat gans on image synthesis},
  author={Dhariwal, Prafulla and Nichol, Alexander},
  journal={Advances in neural information processing systems},
  volume={34},
  pages={8780--8794},
  year={2021}
}

@article{lu2022dpm,
  title={Dpm-solver: A fast ode solver for diffusion probabilistic model sampling in around 10 steps},
  author={Lu, Cheng and Zhou, Yuhao and Bao, Fan and Chen, Jianfei and Li, Chongxuan and Zhu, Jun},
  journal={Advances in Neural Information Processing Systems},
  volume={35},
  pages={5775--5787},
  year={2022}
}

@article{lu2022dpm++,
  title={Dpm-solver++: Fast solver for guided sampling of diffusion probabilistic models},
  author={Lu, Cheng and Zhou, Yuhao and Bao, Fan and Chen, Jianfei and Li, Chongxuan and Zhu, Jun},
  journal={arXiv preprint arXiv:2211.01095},
  year={2022}
}

@article{lu2024simplifying,
  title={Simplifying, stabilizing and scaling continuous-time consistency models},
  author={Lu, Cheng and Song, Yang},
  journal={arXiv preprint arXiv:2410.11081},
  year={2024}
}

@article{brock2018large,
  title={Large Scale GAN Training for High Fidelity Natural Image Synthesis},
  author={Brock, Andrew},
  journal={arXiv preprint arXiv:1809.11096},
  year={2018}
}

@inproceedings{sauer2022stylegan,
  title={Stylegan-xl: Scaling stylegan to large diverse datasets},
  author={Sauer, Axel and Schwarz, Katja and Geiger, Andreas},
  booktitle={ACM SIGGRAPH 2022 conference proceedings},
  pages={1--10},
  year={2022}
}

@inproceedings{karras2024analyzing,
  title={Analyzing and improving the training dynamics of diffusion models},
  author={Karras, Tero and Aittala, Miika and Lehtinen, Jaakko and Hellsten, Janne and Aila, Timo and Laine, Samuli},
  booktitle={Proceedings of the IEEE/CVF Conference on Computer Vision and Pattern Recognition},
  pages={24174--24184},
  year={2024}
}

@article{ma2024sit,
  title={Sit: Exploring flow and diffusion-based generative models with scalable interpolant transformers},
  author={Ma, Nanye and Goldstein, Mark and Albergo, Michael S and Boffi, Nicholas M and Vanden-Eijnden, Eric and Xie, Saining},
  journal={arXiv preprint arXiv:2401.08740},
  year={2024}
}

@misc{alpha2024large,
  title={Large-DiT-ImageNet},
  author={Alpha-VLLM},
  year={2024},
  howpublished={\url{https://github.com/Alpha-VLLM/LLaMA2-Accessory/tree/main/Large-DiT-ImageNet}},
  note={}
}

@inproceedings{peebles2023scalable,
  title={Scalable diffusion models with transformers},
  author={Peebles, William and Xie, Saining},
  booktitle={Proceedings of the IEEE/CVF International Conference on Computer Vision},
  pages={4195--4205},
  year={2023}
}

@inproceedings{bao2023all,
  title={All are worth words: A vit backbone for diffusion models},
  author={Bao, Fan and Nie, Shen and Xue, Kaiwen and Cao, Yue and Li, Chongxuan and Su, Hang and Zhu, Jun},
  booktitle={Proceedings of the IEEE/CVF conference on computer vision and pattern recognition},
  pages={22669--22679},
  year={2023}
}

@inproceedings{hoogeboom2023simple,
  title={simple diffusion: End-to-end diffusion for high resolution images},
  author={Hoogeboom, Emiel and Heek, Jonathan and Salimans, Tim},
  booktitle={International Conference on Machine Learning},
  pages={13213--13232},
  year={2023},
  organization={PMLR}
}

@article{kingma2024understanding,
  title={Understanding diffusion objectives as the elbo with simple data augmentation},
  author={Kingma, Diederik and Gao, Ruiqi},
  journal={Advances in Neural Information Processing Systems},
  volume={36},
  year={2024}
}

@article{tian2024visual,
  title={Visual autoregressive modeling: Scalable image generation via next-scale prediction},
  author={Tian, Keyu and Jiang, Yi and Yuan, Zehuan and Peng, Bingyue and Wang, Liwei},
  journal={arXiv preprint arXiv:2404.02905},
  year={2024}
}

@inproceedings{yan2024diffusion,
  title={Diffusion models without attention},
  author={Yan, Jing Nathan and Gu, Jiatao and Rush, Alexander M},
  booktitle={Proceedings of the IEEE/CVF Conference on Computer Vision and Pattern Recognition},
  pages={8239--8249},
  year={2024}
}

@article{yu2024representation,
  title={Representation alignment for generation: Training diffusion transformers is easier than you think},
  author={Yu, Sihyun and Kwak, Sangkyung and Jang, Huiwon and Jeong, Jongheon and Huang, Jonathan and Shin, Jinwoo and Xie, Saining},
  journal={arXiv preprint arXiv:2410.06940},
  year={2024}
}

@inproceedings{hatamizadeh2025diffit,
  title={Diffit: Diffusion vision transformers for image generation},
  author={Hatamizadeh, Ali and Song, Jiaming and Liu, Guilin and Kautz, Jan and Vahdat, Arash},
  booktitle={European Conference on Computer Vision},
  pages={37--55},
  year={2025},
  organization={Springer}
}

@inproceedings{rombach2022high,
  title={High-resolution image synthesis with latent diffusion models},
  author={Rombach, Robin and Blattmann, Andreas and Lorenz, Dominik and Esser, Patrick and Ommer, Bj{\"o}rn},
  booktitle={Proceedings of the IEEE/CVF conference on computer vision and pattern recognition},
  pages={10684--10695},
  year={2022}
}

@article{heusel2017gans,
  title={Gans trained by a two time-scale update rule converge to a local nash equilibrium},
  author={Heusel, Martin and Ramsauer, Hubert and Unterthiner, Thomas and Nessler, Bernhard and Hochreiter, Sepp},
  journal={Advances in neural information processing systems},
  volume={30},
  year={2017}
}

@inproceedings{li2023mage,
  title={Mage: Masked generative encoder to unify representation learning and image synthesis},
  author={Li, Tianhong and Chang, Huiwen and Mishra, Shlok and Zhang, Han and Katabi, Dina and Krishnan, Dilip},
  booktitle={Proceedings of the IEEE/CVF Conference on Computer Vision and Pattern Recognition},
  pages={2142--2152},
  year={2023}
}

@article{bao2021beit,
  title={Beit: Bert pre-training of image transformers},
  author={Bao, Hangbo and Dong, Li and Piao, Songhao and Wei, Furu},
  journal={arXiv preprint arXiv:2106.08254},
  year={2021}
}

@inproceedings{
    austin2023structured,
    title={Structured Denoising Diffusion Models in Discrete State-Spaces},
    author={Jacob Austin and Daniel D. Johnson and Jonathan Ho and Daniel Tarlow and Rianne van den Berg},
    booktitle={Advances in Neural Information Processing Systems},
    year={2021},
}

@inproceedings{campbell2022continuous,
  title={A Continuous Time Framework for Discrete Denoising Models},
  author={Andrew Campbell and Joe Benton and Valentin De Bortoli and Tom Rainforth and George Deligiannidis and A. Doucet},
  booktitle={Advances in Neural Information Processing Systems},
    year={2022},
}

@article{he2022diffusionbert,
  title={Diffusionbert: Improving generative masked language models with diffusion models},
  author={He, Zhengfu and Sun, Tianxiang and Wang, Kuanning and Huang, Xuanjing and Qiu, Xipeng},
  journal={arXiv preprint arXiv:2211.15029},
  year={2022}
}

@inproceedings{
    Sun2022ScorebasedCD,
    title={Score-based Continuous-time Discrete Diffusion Models},
    author={Haoran Sun and Lijun Yu and Bo Dai and Dale Schuurmans and Hanjun Dai},
    booktitle={The Eleventh International Conference on Learning Representations },
    year={2023},
}

@misc{lou2024discrete,
      title={Discrete Diffusion Modeling by Estimating the Ratios of the Data Distribution}, 
      author={Aaron Lou and Chenlin Meng and Stefano Ermon},
      year={2024},
      eprint={2310.16834},
      archivePrefix={arXiv},
      primaryClass={stat.ML}
}

@article{gat2024discrete,
  title={Discrete Flow Matching},
  author={Gat, Itai and Remez, Tal and Shaul, Neta and Kreuk, Felix and Chen, Ricky TQ and Synnaeve, Gabriel and Adi, Yossi and Lipman, Yaron},
  journal={NeurIPS},
  year={2024}
}

@misc{campbell2024generativeflowsdiscretestatespaces,
      title={Generative Flows on Discrete State-Spaces: Enabling Multimodal Flows with Applications to Protein Co-Design}, 
      author={Andrew Campbell and Jason Yim and Regina Barzilay and Tom Rainforth and Tommi Jaakkola},
      year={2024},
      eprint={2402.04997},
      archivePrefix={arXiv},
      primaryClass={stat.ML},
}

@article{chen2023fast,
  title={Fast Sampling via De-randomization for Discrete Diffusion Models},
  author={Chen, Zixiang and Yuan, Huizhuo and Li, Yongqian and Kou, Yiwen and Zhang, Junkai and Gu, Quanquan},
  journal={arXiv preprint arXiv:2312.09193},
  year={2023}
}

@article{Zheng2023ARD,
  title={A Reparameterized Discrete Diffusion Model for Text Generation},
  author={Lin Zheng and Jianbo Yuan and Lei Yu and Lingpeng Kong},
  journal={ArXiv},
  year={2023},
  volume={abs/2302.05737},
}

@article{song2020score,
  title={Score-based generative modeling through stochastic differential equations},
  author={Song, Yang and Sohl-Dickstein, Jascha and Kingma, Diederik P and Kumar, Abhishek and Ermon, Stefano and Poole, Ben},
  journal={arXiv preprint arXiv:2011.13456},
  year={2020}
}

@article{hoogeboom2021argmax,
  title={Argmax flows and multinomial diffusion: Learning categorical distributions},
  author={Hoogeboom, Emiel and Nielsen, Didrik and Jaini, Priyank and Forr{\'e}, Patrick and Welling, Max},
  journal={NeurIPS},
  volume={34},
  pages={12454--12465},
  year={2021}
}

@misc{hu2024maskneed,
      title={[MASK] is All You Need}, 
      author={Vincent Tao Hu and Björn Ommer},
      year={2024},
      eprint={2412.06787},
      archivePrefix={arXiv},
      primaryClass={cs.CV},
      url={https://arxiv.org/abs/2412.06787}, 
}

@article{goodfellow2014generative,
  title={Generative adversarial nets},
  author={Goodfellow, Ian and Pouget-Abadie, Jean and Mirza, Mehdi and Xu, Bing and Warde-Farley, David and Ozair, Sherjil and Courville, Aaron and Bengio, Yoshua},
  journal={Advances in neural information processing systems},
  volume={27},
  year={2014}
}

@article{sun2024autoregressive,
  title={Autoregressive Model Beats Diffusion: Llama for Scalable Image Generation},
  author={Sun, Peize and Jiang, Yi and Chen, Shoufa and Zhang, Shilong and Peng, Bingyue and Luo, Ping and Yuan, Zehuan},
  journal={arXiv preprint arXiv:2406.06525},
  year={2024}
}

@article{kynkaanniemi2024applying,
  title={Applying guidance in a limited interval improves sample and distribution quality in diffusion models},
  author={Kynk{\"a}{\"a}nniemi, Tuomas and Aittala, Miika and Karras, Tero and Laine, Samuli and Aila, Timo and Lehtinen, Jaakko},
  journal={arXiv preprint arXiv:2404.07724},
  year={2024}
}

@article{chen2024deep,
  title={Deep compression autoencoder for efficient high-resolution diffusion models},
  author={Chen, Junyu and Cai, Han and Chen, Junsong and Xie, Enze and Yang, Shang and Tang, Haotian and Li, Muyang and Lu, Yao and Han, Song},
  journal={arXiv preprint arXiv:2410.10733},
  year={2024}
}

@article{sadat2023cads,
  title={CADS: Unleashing the diversity of diffusion models through condition-annealed sampling},
  author={Sadat, Seyedmorteza and Buhmann, Jakob and Bradley, Derek and Hilliges, Otmar and Weber, Romann M},
  journal={arXiv preprint arXiv:2310.17347},
  year={2023}
}

@article{wang2024analysis,
  title={Analysis of classifier-free guidance weight schedulers},
  author={Wang, Xi and Dufour, Nicolas and Andreou, Nefeli and Cani, Marie-Paule and Abrevaya, Victoria Fern{\'a}ndez and Picard, David and Kalogeiton, Vicky},
  journal={arXiv preprint arXiv:2404.13040},
  year={2024}
}

@inproceedings{bai2024meissonic,
  title={Meissonic: Revitalizing masked generative transformers for efficient high-resolution text-to-image synthesis},
  author={Bai, Jinbin and Ye, Tian and Chow, Wei and Song, Enxin and Chen, Qing-Guo and Li, Xiangtai and Dong, Zhen and Zhu, Lei and Yan, Shuicheng},
  booktitle={The Thirteenth International Conference on Learning Representations},
  year={2024}
}

@article{tang2024hart,
  title={Hart: Efficient visual generation with hybrid autoregressive transformer},
  author={Tang, Haotian and Wu, Yecheng and Yang, Shang and Xie, Enze and Chen, Junsong and Chen, Junyu and Zhang, Zhuoyang and Cai, Han and Lu, Yao and Han, Song},
  journal={arXiv preprint arXiv:2410.10812},
  year={2024}
}

@article{shao2024bag,
  title={Bag of Design Choices for Inference of High-Resolution Masked Generative Transformer},
  author={Shao, Shitong and Zhou, Zikai and Ye, Tian and Bai, Lichen and Xu, Zhiqiang and Xie, Zeke},
  journal={arXiv preprint arXiv:2411.10781},
  year={2024}
}

@inproceedings{ni2024adanat,
  title={Adanat: Exploring adaptive policy for token-based image generation},
  author={Ni, Zanlin and Wang, Yulin and Zhou, Renping and Lu, Rui and Guo, Jiayi and Hu, Jinyi and Liu, Zhiyuan and Yao, Yuan and Huang, Gao},
  booktitle={European Conference on Computer Vision},
  pages={302--319},
  year={2024},
  organization={Springer}
}
\bibliographystyle{icml2025}

\newpage
\appendix
\onecolumn
\section{Equivalence of the masking strategies of MaskGIT and MDM}
\label{app:equival_maskgit_mdm}
\label{sec:loss_equivalence}
In this section, we demonstrate that the masking strategies of MaskGIT and MDM are equivalent in expectation. MaskGIT first samples a ratio \( r \) from \( [0,1] \) and then uniformly masks  
\( \lceil N \gamma_r \rceil \) tokens of \( \boldsymbol{x} \) as \( \text{[M]} \). In contrast, for MDM, each token is independently masked as \( \text{[M]} \) with probability \( \gamma_t \).  

First, for MDM, the cross-entropy loss in \cref{eq:mdm_loss} has multiple equivalent forms~\cite{ou2024your}. To facilitate better understanding, we reformulate \cref{eq:mdm_loss} as an expectation over \( t \):  

\begin{equation}
    \mathcal{L}(\boldsymbol{x_0}) = \mathbb{E}_{t \sim U[0,1]}  \mathbb{E}_{q(\boldsymbol{x}_t|\boldsymbol{x}_0)} \left[ \frac{\gamma_t^\prime}{\gamma_t} \sum_{\{i|\boldsymbol{x}_t^i=\text{[M]}\}}  -\log p_{\boldsymbol{\theta}}(\boldsymbol{x}_0^i|\boldsymbol{x}_t) \right].
\label{eq:mdm_loss_mc}
\end{equation}  

As an example, we consider the linear mask schedule, where \( \gamma_t = t \). In this formulation, the forward process involves independently masking each token  
based on a uniformly sampled \( t \).  Under this setting, the loss simplifies to:  

\begin{equation}
    \mathcal{L}(\boldsymbol{x_0}) = \mathbb{E}_{t \sim U[0,1]}  \mathbb{E}_{q(\boldsymbol{x}_t|\boldsymbol{x}_0)} \left[ \frac{1}{t} \sum_{\{i|\boldsymbol{x}_t^i=\text{[M]}\}}  -\log p_{\boldsymbol{\theta}}(\boldsymbol{x}_0^i|\boldsymbol{x}_t) \right].
\label{eq:mdm_loss_linear}
\end{equation}

For MaskGIT, the number of masked tokens \( l \) is sampled  
from a uniform distribution \( U[1, N] \),  
after which \( l \) tokens in \( \boldsymbol{x}_0 \) are randomly masked as \( \text{[M]} \).  
Under this scheme, the loss function can be rewritten as:  

\begin{equation}
    \mathcal{L}(\boldsymbol{x_0}) = \mathbb{E}_{l \sim U[1,N]}  \mathbb{E}_{q(\boldsymbol{x}_l|\boldsymbol{x}_0)} \left[ \frac{1}{\frac{l}{N}} \sum_{\{i|\boldsymbol{x}_l^i=\text{[M]}\}}  -\log p_{\boldsymbol{\theta}}(\boldsymbol{x}_0^i|\boldsymbol{x}_l) \right].
\label{eq:maskgit_app_loss}
\end{equation}  

As shown in \citet{ou2024your}, \cref{eq:maskgit_app_loss} and \cref{eq:mdm_loss_linear} are equivalent in expectation.  
In this paper, we adopt the formulation of \cref{eq:mdm_loss} with an exponential mask schedule as the default setting.

\section{Mask schedules}
\label{app:mask_schedule}

\subsection{Formulations and Illustrations of Mask Schedules}

\begin{figure}[!t]
    \centering
    \includegraphics[width=.7\linewidth]{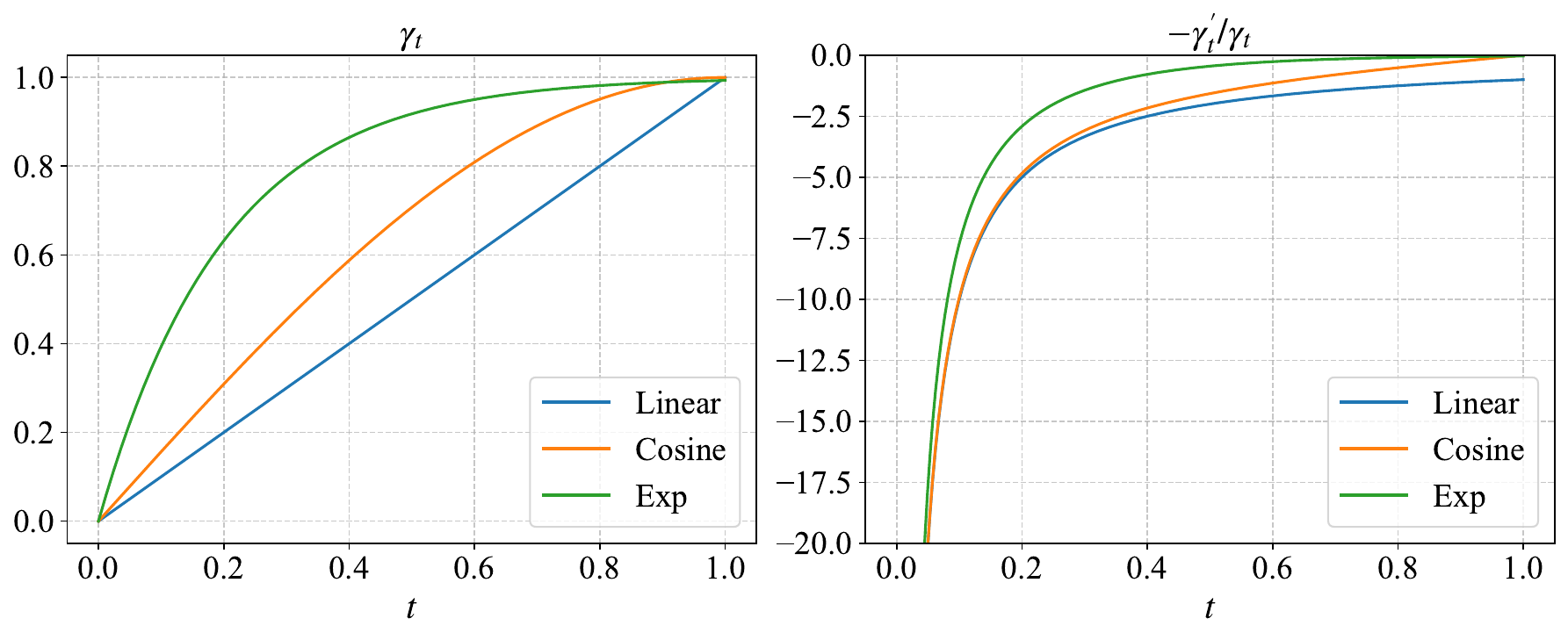}
    \caption{\textbf{Different choices of mask schedules.} Left: $\gamma_t$ (i.e., the probability that each token is masked during the forward process). Right: Weight of the loss in MDM.}
    \label{fig:mask_schedule}
\end{figure}

We present different choices of mask schedules in Fig.~\ref{fig:mask_schedule} and Tab.~\ref{tab:mask_schedule}. The linear schedule achieves the best empirical performance in text generation, as demonstrated in previous work~\cite{loudiscrete, sahoo2024simple, shi2024simplified}. In comparison to the linear schedule, the cosine and exp schedules mask more tokens during the forward process of MDM.

\begin{table}[!t]
    \caption{\label{tab:mask_schedule} \textbf{Mask schedule formulations.}}
    \vskip 0.15in
    \centering
    \begin{tabular}{lcc}
    \toprule
    \textbf{Mask schedule} & $\boldsymbol{\gamma_t}$ & $\boldsymbol{\frac{-\gamma_t'}{\gamma_t}}$ \\
    \midrule
    Linear & $t$ & $-\frac{1}{t}$ \\
    Cosine & $\cos\left(\frac{\pi}{2}(1-t)\right)$ & $-\frac{\pi}{2}\tan\left(\frac{\pi}{2}(1-t)\right)$ \\
    Exp & $1-\exp(-5t)$ & $-\frac{5\exp(-5t)}{1-\exp(-5t)}$ \\
    \bottomrule
    \end{tabular}
\end{table}

\subsection{Sampling Simulator Experiment}
\label{app:sample_similator_exp}
\begin{figure}[!t]
    \centering
    \includegraphics[width=0.5\linewidth]{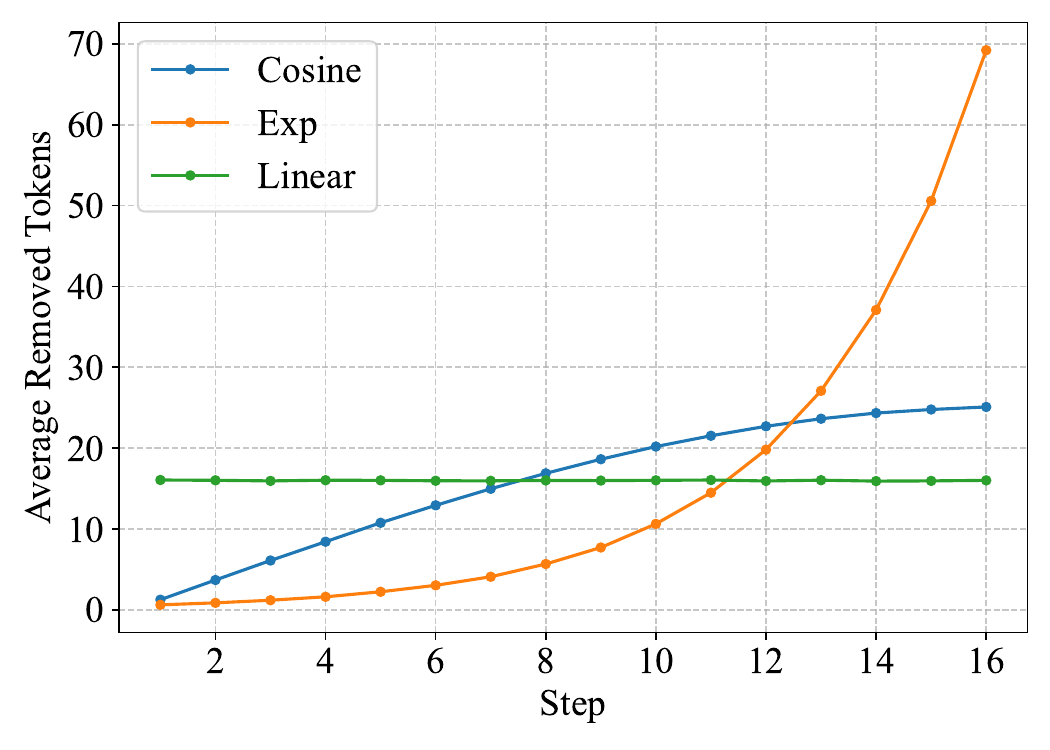}
    \caption{\textbf{Comparison of mask removal} for different sample mask schedule.\label{fig:sample_mask_removal}}
    \label{fig:enter-label}
\end{figure}

During sampling, we conducted a simulation experiment with a total of 256 sample tokens and 16 sampling steps. Therefore, the temporal interval $[0, 1]$ is discretized into $16$ equally sized segments for sampling purposes. Let $s_i$ and $t_i$ represent the endpoint and starting point of the $i$-th segment, respectively, where $i \in \{1, 2, \dots, 16\}$. The indexing is defined such that $t_1$ corresponds to the start of the first segment. Specifically, the endpoints are defined as $s_i = \frac{16 - i}{16}$ and the starting points as $t_i = \frac{16 - i + 1}{16}$. In each step $i$, the prediction for each token is made with a probability of $\frac{\gamma_{t_i} - \gamma_{s_i}}{\gamma_{s_i}}$, as given by \cref{eq:mdm_sample}. We simulated the process 10,000 times and calculated the average number of tokens predicted in each step. The experimental results are shown in Fig.~\ref{fig:sample_mask_removal}. 

We observed the following trends: For the linear schedule, the model predicts almost the same number of tokens in each step. In contrast, for the cosine schedule, the model predicts fewer tokens in the earlier steps and more tokens in the later steps. Compared to the cosine schedule, the exp schedule predicts even fewer tokens in the earlier steps and progressively more tokens in the later steps.

\section{Time Interval for Classifier Free Guidance}
\label{app:time_interval_exp}
\begin{figure}[t]
    \centering
    \subfigure[CFG vs. FID]{
        \includegraphics[width=0.4\linewidth]{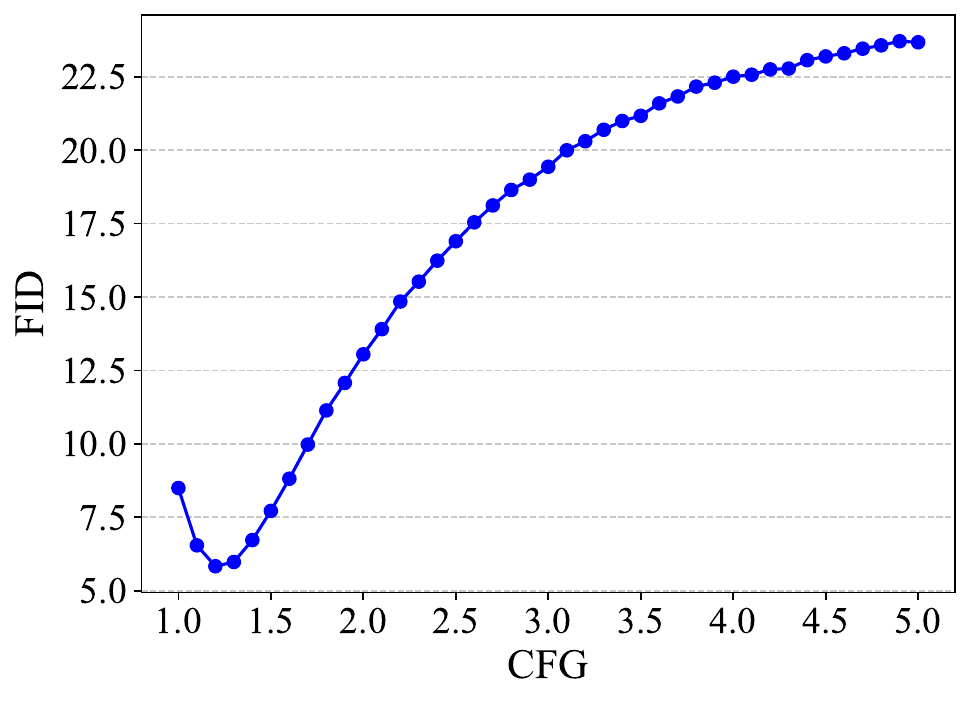}
    }
    \subfigure[CFG vs. IS]{
        \includegraphics[width=0.4\linewidth]{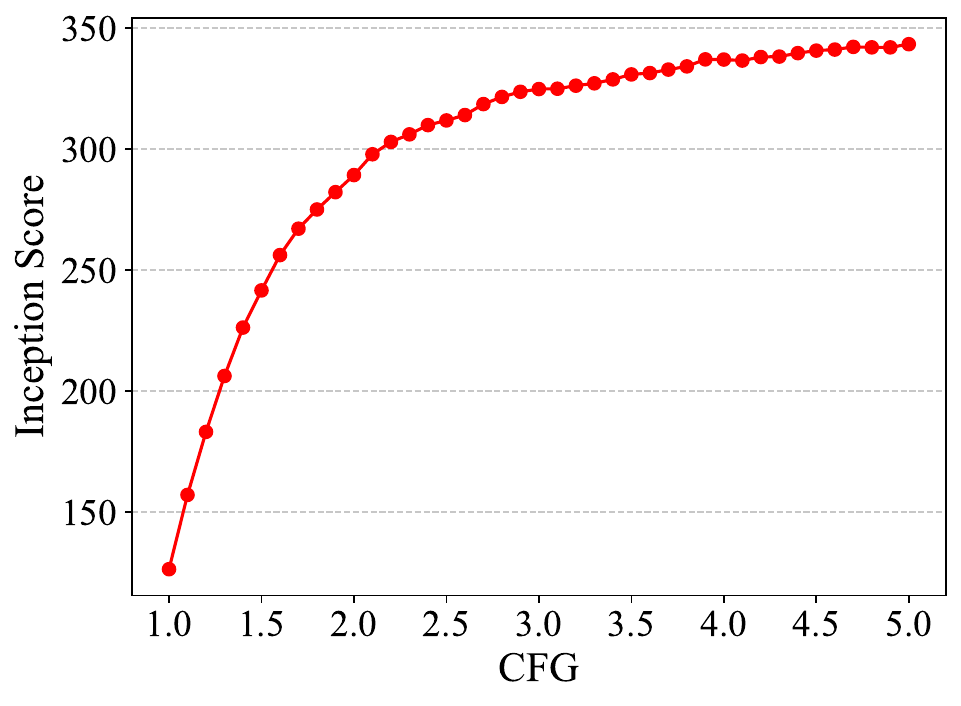}
    }
    \caption{\textbf{Generation performance is sensitive to the CFG value} when using the constant schedule.}
    \label{fig:cfg_ab}
\end{figure}

\begin{figure*}[t!]
    \centering
    \subfigure[FID vs. Time interval\label{fig:time_interval_fid}]{
        \includegraphics[width=0.46\linewidth]{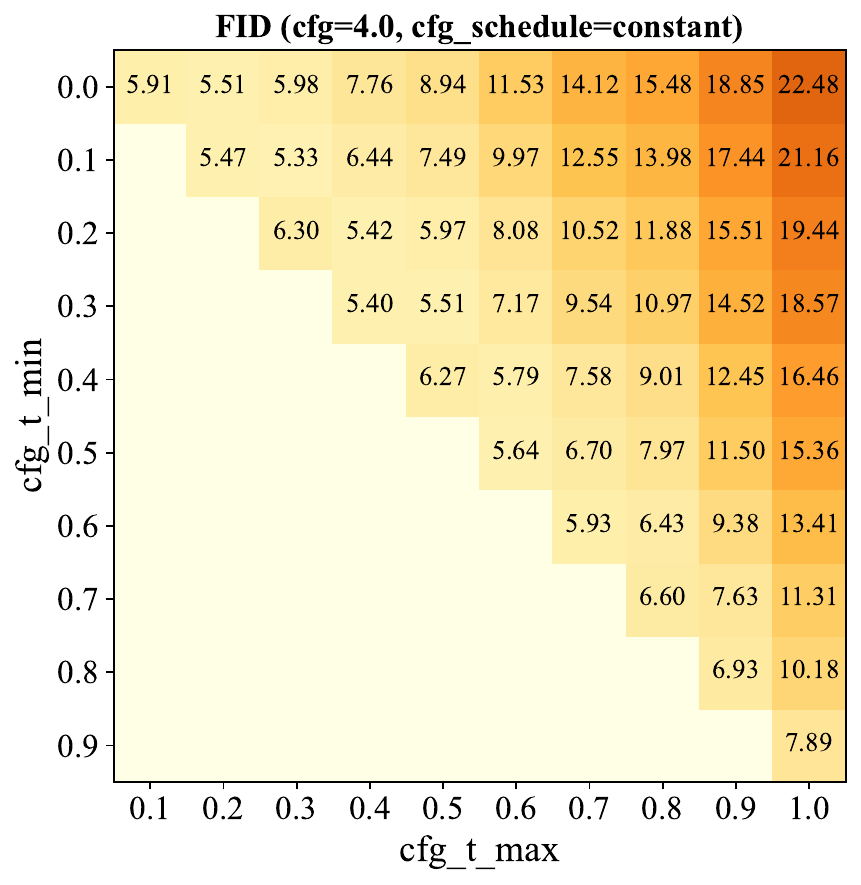}
    }
    \hspace{.1cm}
    \subfigure[IS vs. Time interval\label{fig:time_interval_is}]{
        \includegraphics[width=0.46\linewidth]{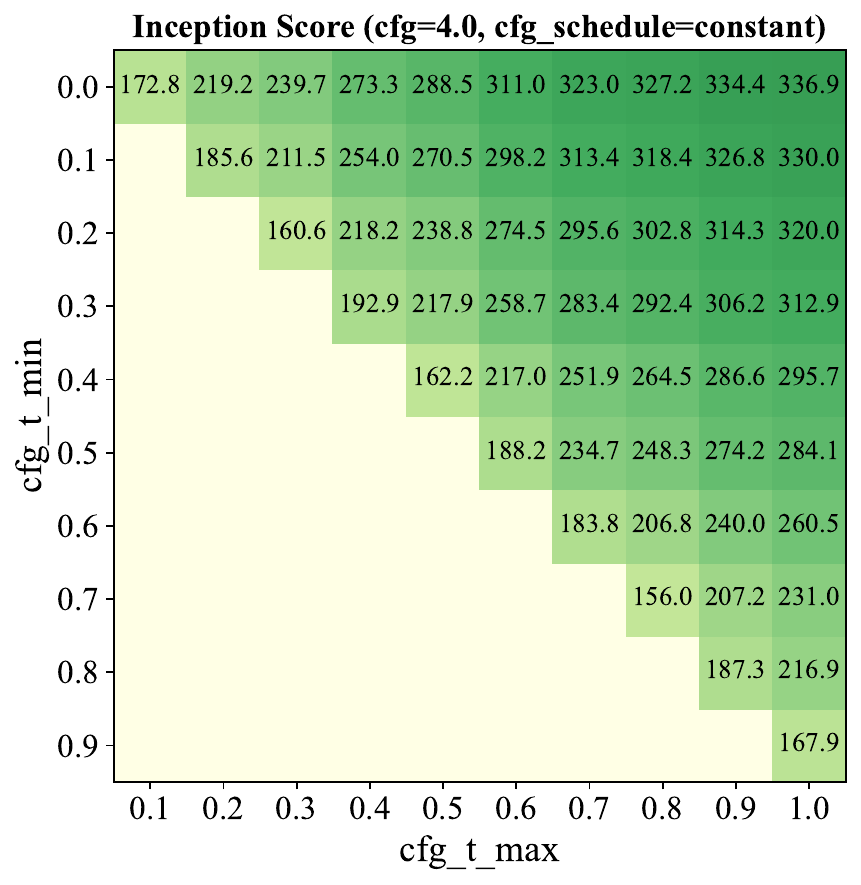}
    }
    \vspace{-1em}
    \caption{\textbf{Performance across different time intervals.} Subplots show (a) FID and (b) Inception Score(IS).}
    \label{fig:time_interval_ab}
\end{figure*}

To validate our hypothesis that an excessively strong guide in the early stages may drastically reduce the variation in generated samples, leading to a higher FID, we conducted an experiment with a total of 256 sample tokens and 16 sampling steps. A more detailed description of the sampling procedure can be found in Appendix~\ref{app:sample_similator_exp}. Let $s_i$ and $t_i$ represent the endpoint and starting point of the $i$-th sampling step, respectively. We define $\text{t}_{\text{min}}$ and $\text{t}_{\text{max}}$ for CFG. If $s_i \in [\text{t}_{\text{min}}, \text{t}_{\text{max}}]$, we apply CFG to guide the sampling; otherwise, we do not use CFG and rely solely on simple conditional generation. As shown in Fig.~\ref{fig:time_interval_ab}, we observe that when $\text{t}_{\text{min}}=0$ and $\text{t}_{\text{max}}=1$, the FID value is 22.48, demonstrating low variation in the generated samples. Additionally, in the top left corner of Fig.~\ref{fig:time_interval_fid} (i.e., when $\text{t}_{\text{min}} < \text{t}_{\text{max}} \leq 0.5$), we achieve a relatively low FID (indicating higher variation), which supports our hypothesis and encourages the application of CFG guidance only during the later stages of sampling.

\begin{table}[t!]
\caption{\label{tab:code_used_and_license} \textbf{The code links and licenses.}}
\vskip 0.15in
\small
\centering
\begin{tabular}{lccccc}
\toprule
Method & Link  &  License \\
\midrule
MAR & \url{https://github.com/LTH14/mar} &  MIT License \\
DPM-Solver & \url{https://github.com/LuChengTHU/dpm-solver} & MIT License \\
DC-AE & \url{https://github.com/mit-han-lab/efficientvit} &  Apache-2.0 license \\
\bottomrule
\end{tabular}
\end{table}
\section{Experiment settings and results}
\label{app:exp_settings}
\begin{table}[t!]
    \caption{\textbf{Training configurations of models on ImageNet 256$\times$256.}} 
    \label{tab:appendix:setting_in256}
    \vskip 0.15in
    \centering
    \begin{tabular}{lccccc}
        \toprule
        \multicolumn{1}{c}{} & \multicolumn{5}{c}{Model Size} \\
        \cmidrule(lr){2-6}
        & XS & S & B & L & H \\
        \midrule
        
        \multicolumn{6}{l}{\textit{Architecture Configurations}} \\
        \midrule
        Transformer blocks & 20  & 24  & 24  & 32  & 40  \\
        Transformer width  & 448 & 512 & 768 & 1024 & 1280 \\
        MLP blocks         & 3   & 3   & 6   & 8    & 12   \\
        MLP width          & 1024 & 1024 & 1024 & 1280 & 1536 \\
        Params (M)         & 69  & 97  & 208 & 478  & 942  \\
        
        \midrule
        
        \multicolumn{6}{l}{\textit{Training Hyperparameters}} \\
        \midrule
        Epochs             &  800   & 800   &800   & 800 & 800 \\ 
        Learning rate      & 4.0e-4 & 4.0e-4 & 8.0e-4 & 8.0e-4 & 8.0e-4 \\
        Batch size         & 1024 & 1024 & 2048 & 2048 & 2048 \\
        Adam $\beta_1$     & 0.9 & 0.9 & 0.9 & 0.9 & 0.9 \\ 
        Adam $\beta_2$     & 0.95 & 0.95 & 0.95 & 0.95  & 0.95\\ 
        \bottomrule
    \end{tabular}
\end{table}

\begin{table}[t!]
    \caption{\textbf{Training configurations of models on ImageNet 512$\times$512.}} 
    \label{tab:appendix:setting_in512}
    \vskip 0.15in
    \centering
    \begin{tabular}{lcccc}
        \toprule
        \multicolumn{1}{c}{} & \multicolumn{4}{c}{Model Size} \\
        \cmidrule(lr){2-5}
        & XS & S & B & L \\
        \midrule
        
        \multicolumn{5}{l}{\textit{Architecture Configurations}} \\
        \midrule
        Transformer blocks & 20  & 24  & 24  & 32  \\
        Transformer width  & 448 & 512 & 768 & 1024 \\
        MLP blocks         & 6   & 6   & 8   & 8    \\
        MLP width          & 1280   & 1280   &1280   & 1280 \\ 
        Params (M)         & 104 & 132 & 244 & 478  \\
        
        \midrule
        
        \multicolumn{5}{l}{\textit{Training Hyperparameters}} \\
        \midrule
        Epochs             &  800   & 800   &800   & 800 \\ 
        Learning rate      & 4.0e-4 & 4.0e-4 & 8.0e-4 & 8.0e-4 \\
        Batch size         & 1024 & 1024 & 2048 & 2048 \\
        Adam $\beta_1$     & 0.9 & 0.9 & 0.9 & 0.9 \\ 
        Adam $\beta_2$     & 0.95 & 0.95 & 0.95 & 0.95 \\ 
        \bottomrule
    \end{tabular}
\end{table}

We implement eMIGM upon the official code of MAR~\cite{li2024autoregressive}, DC-AE~\cite{chen2024deep}, DPM-Solver~\cite{lu2022dpm, lu2022dpm++}, whose code links and licenses are presented in Tab.~\ref{tab:code_used_and_license}.

\textbf{Image Tokenizer.} For ImageNet $256\times256$, we use the same KL-16 image tokenizer as in MAR~\cite{li2024autoregressive}, which has a stride of 16. That is, for an image of size $256\times256$, it outputs an image token sequence of length $16\times16$, with each token having a dimensionality of 16. For ImageNet $512\times512$, we use the DC-AE-f32 tokenizer~\cite{chen2024deep} for efficiency, which has a stride of 32, and each token has a dimensionality of 32.

\textbf{Classifier-Free Guidance (CFG).} In the original CFG, during training, the class condition is replaced with a fake class token with a probability of 10\%. During sampling, the prediction model takes both the class token and the fake class token as input, generating outputs $z_c$ and $z_u$. Conceptually, CFG encourages the generated image to align more closely with the result conditioned on $z_c$ while deviating from the result conditioned on $z_u$. For CFG with Mask, we replace the fake class token with a masked token as the input for unconditional generation. We use a constant CFG schedule and the time interval strategy in our main results presented in Tab.~\ref{tab:in256} and Tab.~\ref{tab:in512}, achieving excellent performance while significantly reducing the sampling cost. Moreover, we observed that with the time interval strategy, we can use a consistently high CFG value to guide generation at each prediction step, eliminating the need for CFG value sweeping.

\textbf{Training Settings.} The detailed training settings for ImageNet $256\times256$ and ImageNet $512\times512$ are provided in Tab.~\ref{tab:appendix:setting_in256} and Tab.~\ref{tab:appendix:setting_in512}, respectively.

\textbf{More Evaluation Metrics.} We present additional evaluation metrics on ImageNet $256 \times 256$ in Tab.~\ref{tab:more_evaluaiton_metrics}.

\begin{table}[t!]
  \centering
  \caption{\textbf{Image generation results on ImageNet $256\times256$.}}
  \label{tab:more_evaluaiton_metrics} 
  \vspace{5pt} 
  \begin{tabular}{@{}l|c|cc|ccc@{}}
    \toprule
    \textbf{Method} & \textbf{NFE} & \textbf{FID$\downarrow$} & \textbf{sFID$\downarrow$} & \textbf{IS$\uparrow$} & \textbf{Precision$\uparrow$} & \textbf{Recall$\uparrow$} \\
    \midrule
    VAR-d30~\cite{tian2024visual} & 10$\times$2 & 1.92 & - & 323.1  & 0.82 & 0.59 \\
    \midrule
    REPA~\cite{yu2024representation} & 250$\times$1.7  & 1.42 & 4.70 & 305.7  & 0.80 & 0.65 \\
    \midrule
    eMIGM-XS  & 16$\times$1.2 & 4.23 & 5.74 & 218.63 & 0.79 & 0.50 \\
    eMIGM-S   & 16$\times$1.2 & 3.44 & 5.31 & 244.16 & 0.80 & 0.53 \\
    eMIGM-B   & 16$\times$1.2 & 2.79 & 5.20 & 284.62 & 0.82 & 0.54 \\
    eMIGM-L   & 16$\times$1.2 & 2.22 & 4.80 & 291.62 & 0.80 & 0.59 \\
    eMIGM-H   & 16$\times$1.2 & 2.02 & 4.66 & 299.36 & 0.80 & 0.60 \\
    \midrule
    eMIGM-XS  & 128$\times$1.4& 3.62 & 5.47 & 224.91 & 0.80 & 0.51 \\
    eMIGM-S   & 128$\times$1.4& 2.87 & 5.53 & 254.48 & 0.80 & 0.54 \\
    eMIGM-B   & 128$\times$1.35&2.32 & 4.63 & 278.97 & 0.81 & 0.57 \\
    eMIGM-L   & 128$\times$1.4& 1.72 & 4.63 & 304.16 & 0.80 & 0.60 \\
    eMIGM-H   & 128$\times$1.4& 1.57 & 4.68 & 305.99 & 0.80 & 0.63 \\
    \bottomrule
  \end{tabular}
  
\end{table}

\textbf{More Mask Schedules.} In this paper, we explored three mask schedules: (1) \emph{Linear}: $\gamma_t = t$; (2) \emph{Cosine}: $\gamma_t = \cos\left(\frac{\pi}{2}(1-t)\right)$; and (3) \emph{Exp}: $\gamma_t = 1 - \exp(-5t)$. All these schedules are designed to satisfy the approximate boundary conditions $\gamma_0 \approx 0$ and $\gamma_1 \approx 1$. We observed that the exp mask schedule, when used in conjunction with $w(t)=1$, achieves superior performance compared to other settings.

Furthermore, we developed a log-exp schedule, $\gamma_t=\frac{\log\left(1+(e^5-1)\cdot t\right)}{5}$, which aims to balance mask ratios by reducing extremes in both high and low masking. Following the experimental setup detailed in Fig.~\ref{fig:fid_loss_weight}, we present the FID results in Tab.~\ref{tab:abla_mask_schedule}. We observed that the log-exp schedule demonstrates improved convergence and performance, thereby validating the benefit of exploring new masking schedules. We leave further investigation of more mask schedules for future work.

\begin{table}[t!]
  \centering
  \caption{\textbf{Ablation study} on different mask schedules, reporting FID scores.}
  \label{tab:abla_mask_schedule}
  \vspace{5pt}
  \begin{tabular}{@{}c|cccc@{}}
    \toprule
    \textbf{Epoch} & \textbf{Linear} & \textbf{Cosine} & \textbf{Exp} & \textbf{Log-Exp} \\
    \midrule
    100 & 38.66 & 24.99 & 28.63 & 25.38 \\
    200 & 30.55 & 16.70 & 17.97 & 11.81 \\
    300 & 24.55 & 15.00 & 11.57 & 12.48 \\
    400 & 24.96 & 12.39 & 11.90 & 9.91  \\
    \bottomrule
  \end{tabular}
\end{table}

\textbf{Sampling Speed Comparison with EDM2.} Compared with EDM2's generation network, EDM2's guidance network is relatively small. We therefore conducted additional experiments to compare sampling speeds on a single A100 GPU (batch size 256), with the results presented in Tab.~\ref{tab:sampling_speed_comparison}. eMIGM-L achieves faster sampling than EDM2-XXL, primarily due to its lower parameter count. Despite requiring a higher NFE, it still maintains competitive performance.

\begin{table}[t!]
  \centering
  \caption{\textbf{Comparison of sampling speed.}} 
  \label{tab:sampling_speed_comparison}
  \vspace{5pt}
  \begin{tabular}{@{}lcc@{}}
    \toprule
    \textbf{Model} & \textbf{Avg sec per image$\downarrow$} & \textbf{FID$\downarrow$} \\
    \midrule
    eMIGM-L & 0.165 & 1.77 \\
    EDM2-XXL~\cite{karras2024analyzing} & 0.552 & 1.81 \\
    EDM2-XXL with guidance interval~\cite{kynkaanniemi2024applying} & 0.481 & 1.40 \\
    \bottomrule
  \end{tabular}
\end{table}


\end{document}